\documentclass[journal]{IEEEtran}
\usepackage{balance}

\usepackage[T1]{fontenc}
\usepackage[utf8]{inputenc}

\usepackage{amsmath}
\usepackage{amssymb}

\usepackage{graphicx}
\usepackage{svg}

\usepackage{algorithm}
\usepackage[noend]{algpseudocode}

\usepackage{booktabs}
\usepackage{multirow}
\usepackage{makecell}
\usepackage{colortbl}
\usepackage{arydshln}

\usepackage[normalem]{ulem}
\usepackage{enumitem}
\usepackage{xcolor}
\usepackage{listings}

\usepackage{academicons}

\newcommand{\orcidicon}{%
    \includegraphics[width=8pt]{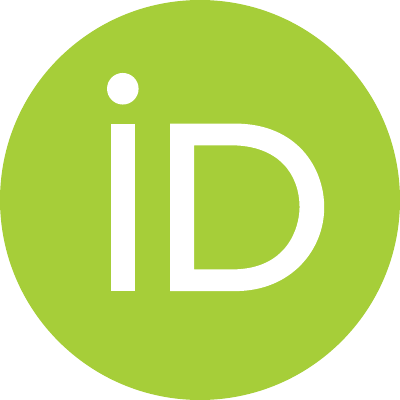}%
}

\usepackage{hyperref}

\algrenewcommand{\algorithmiccomment}[1]{%
  \hfill{\footnotesize\color{black!55}$\triangleright$~#1}%
}

\ifCLASSINFOpdf
\else
\fi
\begin{document}
%

\title{{\textit{ShadowCast}:} \\On-the-Fly Context-Aware Shadow Augmentation \\for Remote Sensing Image Segmentation \vspace{-0.2cm}}

%
%
%



\author{\IEEEauthorblockN{Dongho Yoon},
\IEEEauthorblockN{Yechan Kim\textsuperscript{\href{https://orcid.org/0000-0002-2438-3590}{\orcidicon{}}}}, 
\IEEEauthorblockN{Jongmin Joo}, 
\IEEEauthorblockN{JongHyun Park}, and 
\IEEEauthorblockN{Moongu Jeon\textsuperscript{\href{https://orcid.org/0000-0002-2775-7789}{\orcidicon{}}}}
\vspace{-0.8cm}
\thanks{Received \textcolor{black}{28 August 2026}; revised \textcolor{black}{21 October 2026}; accepted \textcolor{black}{00 00000000 2026}; Date of publication \textcolor{black}{00 00000000 2026}; date of current version \textcolor{black}{00 00000000 2026}. \textcolor{black}{This work was partly supported by the National Research Foundation of Korea (NRF) grant funded by the Korea government (MSIT) (No. RS-2026-25475324), and GIST SCENT - HPC AI (A100). \textcolor{black}{(\textit{Corresponding authors}: \textit{Moongu Jeon; Yechan Kim.}) }} }
\thanks{D. Yoon, J. Joo, J. Park and M. Jeon are with Gwangju Institute of Science and Technology (GIST), 123 Cheomdangwagi-ro, Buk-gu, Jeonnam-Gwangju, 61005, Korea (e-mail: 
    \href{mailto:gidong76@gm.gist.ac.kr}{gidong76@gm.gist.ac.kr},
    \href{mailto:luck2u99@gm.gist.ac.kr}{luck2u99@gm.gist.ac.kr},
    \href{mailto:citizen135@gm.gist.ac.kr}{citizen135@gm.gist.ac.kr},
    \href{mailto:mgjeon@gist.ac.kr}{mgjeon@gist.ac.kr}).}
\thanks{Y. Kim is with AI Laboratory, LIG Defense\&Aerospace, 333 Pangyo-ro, Bundang-gu, Seongnam-si, Gyeonggi-do, 13488, Korea (e-mail: 
    \href{mailto:yechan.kim26@ligdna.com}{yechan.kim26@ligdna.com}).
}
\thanks{This article has supplementary downloadable material available at \url{https://doi.org/10.1109/LGRS.2026.xxxxxxxx}, provided by the authors.}
\thanks{Digital Object Identifier 10.1109/LGRS.2026.xxxxxxx}}

%
%

\markboth{}%
{Shell \MakeLowercase{\textit{et al.}}: Bare Demo of IEEEtran.cls for Journals}
%



\maketitle

\begin{abstract}
Shadows cast by elevated objects alter the spectral appearance and obscure the boundaries of objects in remote sensing images, thereby degrading semantic segmentation performance.
    Most existing approaches handle shadows as degradations to be detected and removed, requiring additional costs from dedicated networks and shadow-specific annotations.
We propose \textbf{\textit{ShadowCast}}, which takes the opposite approach: it synthesizes shadows and injects them during training for remote sensing segmentation.
    It extracts masks of elevated objects, such as buildings and trees, directly from the segmentation labels and estimates an image-specific cast-shadow direction cue by probing the brightness around them.
It then projects synthetic shadows along the inferred cast-shadow direction, requiring no learned network and no acquisition metadata.
     Experiments on three widely used remote sensing benchmarks, Vaihingen, Potsdam, and LoveDA, show that \textbf{\textit{ShadowCast}} consistently outperforms the baseline across diverse state-of-the-art segmentation architectures.
Moreover, it brings additional performance gains when combined with standard augmentation and regularization techniques, indicating that stochastic shadow injection serves as a simple and complementary augmentation strategy for remote sensing segmentation.


\end{abstract}

\begin{IEEEkeywords}
Data augmentation, shadow injection, remote sensing image segmentation.
\end{IEEEkeywords}

%
\IEEEpeerreviewmaketitle

\vspace{-0.2cm}
\section{Introduction}

\IEEEPARstart{R}{emote} sensing (RS) image segmentation is fundamental to applications such as urban planning, land-cover mapping, and disaster response. 
    However, its performance is affected by substantial variations in object scale, spectral appearance, spatial context, and imaging conditions. 
Among these factors, this work focuses on cast shadows generated by elevated objects such as buildings and trees. 
    Their location and extent vary with illumination geometry and scene structure \cite{masquil2025s}, while reduced illumination alters the spectral appearance of the underlying surfaces and can obscure object boundaries \cite{liu2023shadow,yu2024enhancing}. 
Although the semantic identity of a surface remains unchanged, its illuminated and shadowed regions may exhibit substantially different visual characteristics.
    Consequently, cast shadows can increase intra-class appearance variation and inter-class spectral confusion, thereby degrading segmentation predictions in shadowed regions.

Most shadow studies for RS image understanding have formulated the problem primarily as shadow detection, removal, or radiometric reconstruction \cite{singh2012shadow,zhang2014object,wang2023shadow,wang2024recreating, shao2024synthesis,shao2025generative,zhang2025shadow,chi2025cross, masquil2025s,bo2026saru}.
    Although such approaches are intuitive, they commonly rely on multi-stage pipelines involving dedicated shadow detectors and restoration modules, resulting in additional computational and annotation costs.
Besides, errors in shadow detection can propagate to subsequent restoration and segmentation stages, while imperfect restoration may alter non-shadow regions or suppress task-relevant spectral and structural cues.
A natural alternative is to expose the segmentation model to shadow-induced appearance variations during training.
    However, generic shadow augmentation, such as RandomShadow and PlasmaShadow \cite{buslaev2020albumentations}, darkens randomly generated image regions without considering which scene objects can plausibly cast shadows.


To address this challenge, we propose \textbf{\textit{ShadowCast}}, a learning-free, context-aware augmentation method that conditions shadow synthesis on information already available in each image-label pair.
    Here, \emph{context-aware} refers to conditioning the augmentation on two complementary sources of scene context.
First, \emph{semantic context} from the segmentation labels identifies elevated objects, such as buildings and trees, that can plausibly cast shadows.
    Second, \emph{photometric context} from the input image provides an image-specific cast-shadow direction cue through the local brightness distribution around these objects.
Using these two cues, \textbf{\textit{ShadowCast}} projects synthetic shadows from plausible shadow-casting objects along the inferred cast-shadow direction.
    Because the entire procedure relies only on the input image and its segmentation label, it does not require additional detection/generative network, shadow-specific annotations, and acquisition metadata such as solar azimuth, solar elevation, and sensor geometry.


We evaluate \textbf{\textit{ShadowCast}} on three widely used RS semantic segmentation benchmarks: ISPRS Vaihingen, Potsdam, and LoveDA, across diverse segmentation architectures.
    The results consistently demonstrate the effectiveness of context-aware shadow augmentation, with \textbf{\textit{ShadowCast}} outperforming both standard training and existing shadow augmentations.
Moreover, its consistent gains when combined with existing augmentation and regularization strategies show that \textbf{\textit{ShadowCast}} provides a complementary source of training variation.



\vspace{-0.1cm}
\section{Related Work} 


\begin{figure}[!t]
    \centering
    \includegraphics[height=4.05cm]{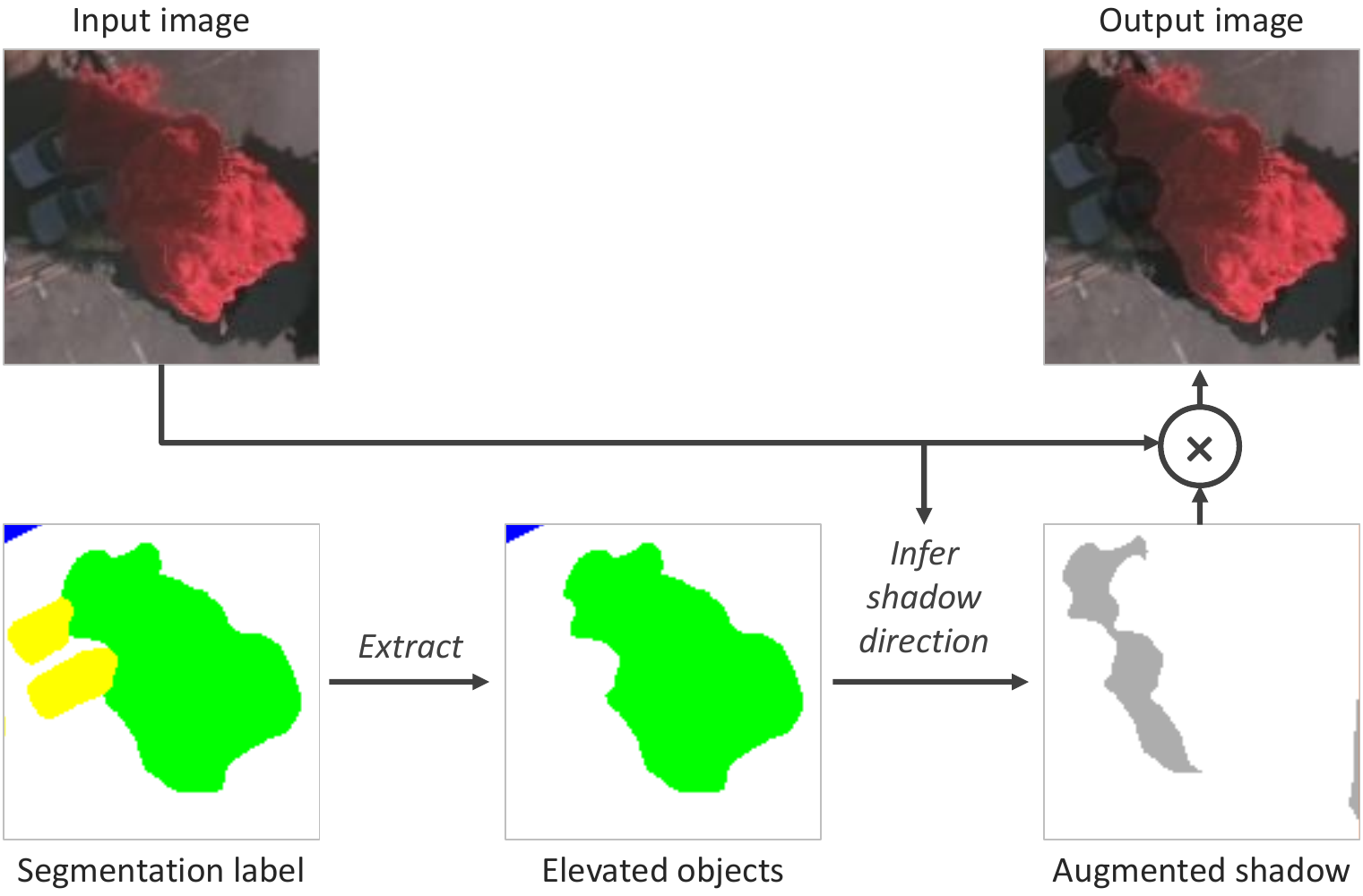} \vspace{-0.2cm}
    \caption{Pipeline of our \textbf{\textit{ShadowCast}} for remote sensing image segmentation.}
    \label{fig1}
    \vspace{-0.4cm}
\end{figure}

Existing shadow processing methods in RS imagery have primarily focused on shadow detection and removal. Early studies employed spectral indices, morphological operations \cite{singh2012shadow}, and object-level spatial relationships \cite{zhang2014object}. Subsequent approaches incorporated spectral reconstruction for complex terrain \cite{wang2023shadow}, multi-spectral radiance characteristics \cite{liu2023shadow}, and geometry-aware illumination compensation \cite{zhang2025shadow}. More recently, Bo et al. \cite{bo2026saru} unified shadow detection and removal, while S-EO \cite{masquil2025s} exploited solar geometry and LiDAR-derived surface models for physically consistent shadow modeling. However, these approaches primarily aim to recover shadow-free imagery and may require dedicated models, shadow annotations, or additional acquisition information.

Beyond shadow processing, various data augmentation methods have been explored to improve segmentation generalization. Conventional approaches, such as CutOut \cite{devries2017improved}, CutMix \cite{yun2019cutmix}, and Random Erasing \cite{zhong2020random}, introduce training variations without explicitly considering scene semantics. 
More recent RS-specific approaches exploit task-related context to guide augmentation. 
Hu et al. \cite{hu2024domain} employ input-level domain mixing and masked feature augmentation to improve cross-domain generalization.
NSegment \cite{kim2025nsegment} applies label-only deformations to improve robustness against annotation inconsistencies.
Li et al. \cite{li2026geospatial} extract geographic units and their spatial relationships from annotations to augment samples of easily confused classes. 
    Unlike these methods, \textbf{\textit{ShadowCast}} synthesizes cast-shadow-induced appearance variations by combining semantic labels with photometric image context to improve segmentation robustness to shadows commonly encountered in satellite and aerial imagery, without requiring additional models, shadow labels, or acquisition metadata.

\vspace{-0.2cm}
\section{Proposed Method}
\vspace{-0.1cm}


\subsection{Motivation}

We build on two observations. 
    First, segmentation labels already provide the spatial extent of elevated objects, such as buildings and trees, which can serve as candidate shadow-casting regions without additional annotation. 
Second, since objects within a single RS image are typically captured under a common acquisition illumination, their cast shadows tend to extend in a dominant direction that is reflected in the local brightness distribution around these objects.
    These properties make it possible to synthesize geometrically plausible cast shadows directly from the input image and its segmentation labels at low cost without auxiliary models. 
Based on these insights, we propose \textit{\textbf{ShadowCast}}, which extracts elevated-object masks from the labels and estimates an image-specific cast-shadow direction cue from local appearance.

\vspace{-0.2cm}
\subsection{Method}
Given an image--label pair $(\mathbf{I},\mathbf{L})$, the proposed \textbf{\textit{ShadowCast}} method generates an augmented image $\mathbf{I}^{\star}$ by projecting the label-derived masks of elevated objects along an image-specific cast-shadow direction.
The semantic mask label $\mathbf{L}$ remains unchanged as the transformation modifies only the illumination of the underlying surfaces without changing their semantic identities.
Note that the overall procedure is summarized in Algorithm~\ref{alg:shadowcast} and Fig.~\ref{fig1}.

\subsubsection{Elevated Object Region (or Mask) Extraction}

Let $\mathcal{E}$ denote the set of semantic classes corresponding to elevated objects that can cast shadows, such as buildings and trees.
We construct a binary elevated-object mask as
\begin{equation}
\mathbf{M}
=
\mathbf{1} \left[\mathbf{L}\in\mathcal{E}\right],
\label{eq1}
\end{equation}
where $\mathbf{1}[\cdot]$ is the indicator function applied pixel-wise.
Thus, $\mathbf{M}(\mathbf{x})=1$ when the label at pixel coordinate $\mathbf{x}$ belongs to one of elevated object classes and $\mathbf{M}(\mathbf{x})=0$ otherwise.
As $\mathbf{M}$ is obtained directly from the semantic segmentation masks in existing RS datasets, no additional object detector, shadow annotation, or geometric metadata is required in our algorithm.

The transformation is applied with the user-defined probability $p$.
Specifically, we sample $u\sim\mathcal{U}(0,1)$ and retain the original image when $u>p$.
We also skip the transformation when no elevated-class pixel is present, i.e., when $\operatorname{sum}(\mathbf{M})=0$.
These conditions avoid unnecessary computation.

\subsubsection{Image-Specific Cast-Shadow Direction Estimation}

Because objects within an aerial or satellite image are commonly captured under the same acquisition illumination, their cast shadows tend to extend in a dominant direction.
We estimate this cast-shadow direction cue by comparing the brightness of regions surrounding the elevated-object mask.

First, the input image $\mathbf{I}$ is converted into a single-channel intensity image as follows:
\begin{equation}
\mathbf{I}'=\operatorname{toGray}(\mathbf{I}),
\label{eq2}
\end{equation}
where $\operatorname{toGray}(\cdot)$ denotes a standard grayscale conversion.
We then define a discrete set of eight candidate directions,
\begin{equation}
\Theta
=
\{(45k)^\circ \mid k=0,\cdots,7\}.
\label{eq3}
\end{equation}
These directions uniformly cover the full angular range at intervals of $45^{\circ}$, providing a lightweight approximation of the dominant cast-shadow orientation.

For a candidate direction $\theta\in\Theta$, we translate the elevated-object mask by a fixed probe distance $r$:
\begin{equation}
\mathbf{R}_{\theta}\gets
\operatorname{shift}\!\left(
\mathbf{M},
\left\lfloor
r[\cos\theta,\sin\theta]^{\mathsf T}
\right\rceil
\right)
\land\neg\mathbf{M},
\quad \forall\theta\in\Theta.
\label{eq4}
\end{equation}
Here, $\operatorname{shift}(\mathbf{M},\mathbf{t})$ translates the binary mask by the two-dimensional integer displacement vector $\mathbf{t}$.
The notation $\lfloor\cdot\rceil$ denotes element-wise rounding to the nearest integer, which converts the continuous displacement determined by $r$, $\cos\theta$, and $\sin\theta$ into valid pixel offsets.
    Note that pixels shifted beyond the image boundary are discarded.
Furthermore, $\land$ and $\neg$ denote element-wise logical conjunction and negation, respectively.
    Hence, the operation $\land\neg\mathbf{M}$ removes the original elevated-object pixels from the translated mask, ensuring that brightness is measured only in neighboring regions rather than on the shadow-casting objects themselves.

\begin{figure}[!t]
    \centering
    \includegraphics[width=4.9cm]{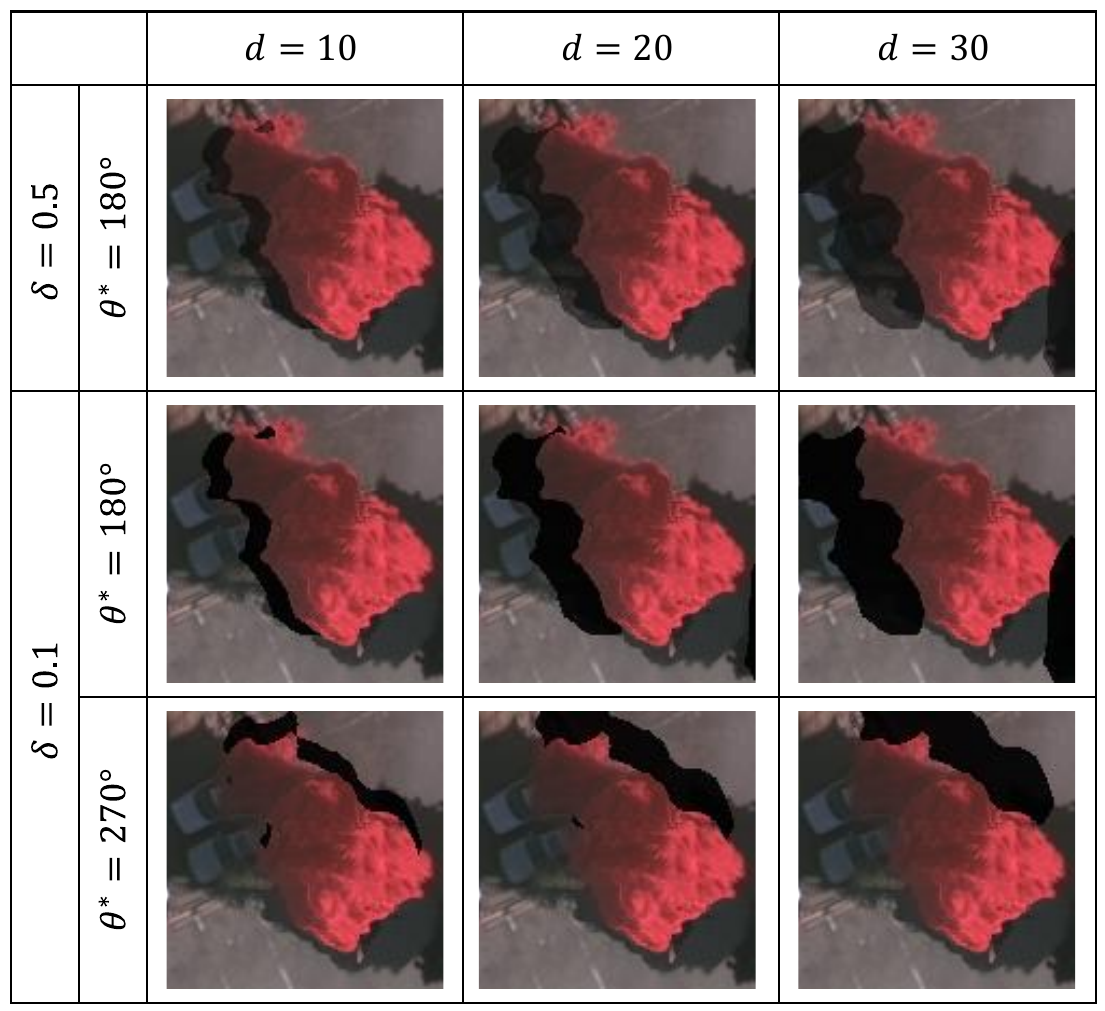} \vspace{-0.2cm}
\caption{Behaviors of \textbf{\textit{ShadowCast}} under varying values of $\theta^{\star}$, $d$, and $\delta$.}
    \label{fig2}
    \vspace{-0.3cm}
\end{figure}

Some shifted probe regions may contain too few valid pixels to support a plausible shadow.
    We therefore define the set of valid candidate directions as
\begin{equation}
\Theta'\gets
\left\{
\theta\in\Theta
\mid
\operatorname{sum}(\mathbf{R}_{\theta})> \tau
\right\},
\label{eq5}
\end{equation}
where $\tau$ denotes a minimum region threshold.
    When $\Theta'$ is empty, the input image is returned without modification.
Otherwise, the mean grayscale intensity of each valid probe region is computed, and the darkest direction is selected:
\begin{equation}
\widetilde{\theta} \gets
\arg\min_{\theta\in\Theta'}
\operatorname{mean}\!\left(
\mathbf{I}'[\mathbf{R}_{\theta}]
\right).
\label{eq6}
\end{equation}
The notation $\mathbf{I}'[\mathbf{R}_{\theta}]$ represents the intensity values at pixels for which $\mathbf{R}_{\theta}=1$.
    In other words, the direction with the lowest average brightness provides a simple estimate of the dominant cast-shadow direction.
This estimation requires neither acquisition metadata nor a learned illumination-estimation model.
    Note that since this brightness-based heuristic can be biased when a probe region overlaps an intrinsically dark surface, such as asphalt or water, the estimated direction should be regarded as a lightweight cast-shadow direction cue rather than an exact recovery of the physical shadow geometry.

To increase augmentation diversity while preserving the estimated cast-shadow directional tendency, we add a bounded angular perturbation as follows:
\begin{equation}
\theta^\star\gets \widetilde{\theta} + \epsilon, \quad  \epsilon \sim \mathcal{U}(-\eta,\eta),
\label{eq7}
\end{equation}
where $\eta$ controls the angular jitter magnitude.

\subsubsection{Synthetic Shadow Projection}

After estimating the cast-shadow direction, we sample a projection distance
\begin{equation}
d\sim\mathcal{U}(d_{\min},d_{\max}),
\label{eq8}
\end{equation}
where $d_{\min}$ and $d_{\max}$ determine the range of synthetic shadow lengths in pixels.
The synthetic shadow mask is then generated by translating the elevated-object mask along $\theta^{\star}$:
\begin{equation}
\mathbf{M}'\gets
\operatorname{shift}(\mathbf{M}, \left\lfloor
d[\cos \theta^\star,
\sin \theta^\star]^{\mathsf T}
\right\rceil)
\land\neg\mathbf{M}.
\label{eq9}
\end{equation}
As in Eq. (\ref{eq4}), the continuous displacement is rounded element-wise to the nearest integer pixel coordinate.
Removing the original mask through $\land\neg\mathbf{M}$ prevents the appearance of the elevated objects themselves from being attenuated.
Therefore, the transformation primarily affects neighboring surfaces onto which a cast shadow may plausibly be projected.

The shadow strength is controlled using a multiplicative attenuation factor $\delta$ as follows:
\begin{equation}
\delta
\sim
\mathcal{U}(\delta_{\min},\delta_{\max}),
\label{eq10}
\end{equation}
where $0<\delta_{\min}\leq\delta_{\max}<1$.
The final augmented image $\mathbf{I}^{\star}$ is defined as in \textcolor{black}{lines 18-19 of Algorithm~\ref{alg:shadowcast}}.
Randomly sampling $\theta^{\star}$, $d$ and $\delta$ increases data diversity in shadow orientation, size, and darkness across training iterations; \textcolor{black}{refer to Fig.~\ref{fig2}}.

\textbf{\textit{ShadowCast}} is fully learning-free, online and operates only on the image and its existing semantic label.
    Its main operations consist of binary-mask translation and masked intensity averaging, making it suitable for online data augmentation.
            For example, on a $512\times512$ Vaihingen training crop, \textbf{\textit{ShadowCast}} requires an average of 16.55 ms per image over 1,000 runs on an Intel$^{\text{(R)}}$ Xeon$^{\text{(R)}}$ Gold 6330 CPU, supporting its use as an on-the-fly augmentation.
Moreover, the local brightness distribution provides an image-conditioned cast-shadow direction cue that is more effective for augmentation than arbitrary direction sampling, as demonstrated in Table~\ref{tab2} and in comparison with prior work; see Tables~\ref{new:shadowaugseries} and \ref{tab4}.

\begin{algorithm}[t]
\caption{Procedure of \textbf{\textit{ShadowCast}}.}
\label{alg:shadowcast}
\begin{algorithmic}[1]

\Require Image--label pair $(\mathbf{I},\mathbf{L})$;
elevated-class set $\mathcal{E}$;
transform probability $p$;
probe distance $r$;
shadow-length range $(d_{\min},d_{\max})$;
angular jitter magnitude $\eta$;
minimum region threshold $\tau$;
brightness attenuation range $(\delta_{\min},\delta_{\max})$

\Ensure Transformed image $\mathbf{I}^{\star}$

\State $u\sim\mathcal{U}(0,1)$
\If{$u>p$}
    \State \Return $\mathbf{I}$
        \Comment{No transformation}
\EndIf

\State $\mathbf{M}\gets\mathbf{1}[\mathbf{L}\in\mathcal{E}]$
\If{$\operatorname{sum}(\mathbf{M})=0$}
    \State \Return $\mathbf{I}$
        \Comment{No transformation}
\EndIf

\State $\mathbf{I}'\gets\operatorname{toGray}(\mathbf{I})$
    \Comment{Enable luminance-based direction selection}
\State $\Theta\gets  \{(45k)^\circ \mid k=0,\cdots,7\}$

\State $\mathbf{R}_{\theta}\gets
\operatorname{shift}\!\left(
\mathbf{M},
\left\lfloor
r[\cos\theta,\sin\theta]^{\mathsf T}
\right\rceil
\right)
\land\neg\mathbf{M},
\quad \forall\theta\in\Theta$
    \Comment{Construct probe regions excluding elevated-object pixels}

\State $\Theta'\gets
\left\{
\theta\in\Theta
\mid
\operatorname{sum}(\mathbf{R}_{\theta})> \tau
\right\}$

\If{$\Theta'=\varnothing$}
    \State \Return $\mathbf{I}$
        \Comment{No transformation}
\EndIf

\State $ \widetilde{\theta} \gets
\arg\min_{\theta\in\Theta'}
\operatorname{mean}\!\left(
\mathbf{I}'[\mathbf{R}_{\theta}]
\right)$
    \Comment{Select the direction $\widetilde{\theta}$ with the darkest probe region}

\State $\theta^\star\gets \widetilde{\theta} + \epsilon, \quad  \epsilon \sim \mathcal{U}(-\eta,\eta)$

\State $d\sim\mathcal{U}(d_{\min},d_{\max})$

\State $\mathbf{M}'\gets
\operatorname{shift}(\mathbf{M}, \left\lfloor
d[\cos \theta^\star,
\sin \theta^\star]^{\mathsf T}
\right\rceil)
\land\neg\mathbf{M}$
\Comment{Extract the target region where the shadow will be applied}

\State $\delta\sim
\mathcal{U}(\delta_{\min},\delta_{\max})$


\State $\mathbf{I}^{\star}\gets\mathbf{I}$ 
\State $\mathbf{I}^{\star}[\mathbf{M}'] \gets\delta\mathbf{I}[\mathbf{M}']$

\State \Return $\mathbf{I}^{\star}$
\Comment{Transformed image by \textbf{\textit{ShadowCast}}}

\end{algorithmic}
\end{algorithm}

\begin{table}[t!]
\vspace{-0.1cm} 
\centering
\caption{Effect of cast-shadow direction selection in \textit{\textbf{ShadowCast}} across three benchmarks
using DeepLabV3+ ($p=0.5$; $d_{\min}=5$; $d_{\max}=30$; $\delta_{\min}=0.5$; $\delta_{\max}=0.8$; $\eta=15$; $\tau=50$; $r=8$).} \vspace{-0.2cm}
\resizebox{6.2cm}{!}{
\setlength{\tabcolsep}{1.5pt}
\renewcommand{\arraystretch}{1.05}
\begin{tabular}{lccc}
\toprule
Cast-shadow direction & Vaihingen & Potsdam & LoveDA \\
\midrule
No shadow augmentation & 77.53$\pm$0.30 & 82.95$\pm$0.10 & 43.29$\pm$0.77 \\
\midrule
Random direction & 78.14$\pm$0.10 & 83.13$\pm$0.14 & 43.61$\pm$0.50\\
\midrule
\cellcolor[HTML]{FFCE93}Estimated direction & \cellcolor[HTML]{FFCE93}78.20$\pm$0.07 & \cellcolor[HTML]{FFCE93}83.22$\pm$0.13 & \cellcolor[HTML]{FFCE93}44.45$\pm$0.30 \\
\cellcolor[HTML]{FFCE93} \;+ Jittering (\textbf{Ours}) & \cellcolor[HTML]{FFCE93}\textbf{78.46$\pm$0.19} & \cellcolor[HTML]{FFCE93}\textbf{83.32$\pm$0.09}  & \cellcolor[HTML]{FFCE93}\textbf{44.89$\pm$0.09}  \\
\bottomrule
\end{tabular}
} \vspace{-0.4cm}
\label{tab2}
\end{table}

\begin{table}[t!]
\vspace{-0.1cm}
\centering
\caption{Impact of \textbf{\textit{ShadowCast}} on training state-of-the-art
segmentation models on Vaihingen, Potsdam, and LoveDA.} \vspace{-0.2cm}

\newcommand{\modelentry}[2]{%
\shortstack[l]{#1\\{\tiny\cite{#2}}}%
}
\newcommand{\basescore}[2]{%
\shortstack[c]{#1$\pm$#2\\{\phantom{(+0.00)}}}%
}
\newcommand{\scscore}[3]{%
\cellcolor[HTML]{FFCE93}%
\shortstack[c]{\textbf{#1$\pm$#2}\\{\textbf{(\textcolor{red}{+#3})}}}%
}

\resizebox{0.96\columnwidth}{!}{%
\setlength{\tabcolsep}{1.5pt}
\renewcommand{\arraystretch}{1.05}
\begin{tabular}{l cc cc cc}
\toprule
\multirow{2}{*}{Model}
& \multicolumn{2}{c}{Vaihingen}
& \multicolumn{2}{c}{Potsdam}
& \multicolumn{2}{c}{LoveDA} \\
\cmidrule(lr){2-3}\cmidrule(lr){4-5}\cmidrule(lr){6-7}
& Baseline & \cellcolor[HTML]{FFCE93}\textbf{\textit{ShadowCast}}
& Baseline & \cellcolor[HTML]{FFCE93}\textbf{\textit{ShadowCast}}
& Baseline & \cellcolor[HTML]{FFCE93}\textbf{\textit{ShadowCast}} \\
\midrule

\modelentry{DeepLabV3+}{deeplabv3plus}
& \basescore{77.53}{0.30} & \scscore{78.46}{0.19}{0.93}
& \basescore{82.95}{0.10} & \scscore{83.32}{0.09}{0.37}
& \basescore{43.29}{0.77} & \scscore{44.89}{0.09}{1.60} \\

\modelentry{DANet}{danet}
& \basescore{79.59}{0.61} & \scscore{80.20}{0.06}{0.61}
& \basescore{84.47}{0.24} & \scscore{84.88}{0.04}{0.41}
& \basescore{43.35}{0.78} & \scscore{44.43}{0.85}{1.08} \\

\modelentry{APCNet}{apcnet}
& \basescore{79.15}{0.85} & \scscore{80.53}{0.18}{1.38}
& \basescore{84.73}{0.13} & \scscore{85.05}{0.01}{0.32}
& \basescore{46.46}{0.60} & \scscore{47.94}{0.17}{1.48} \\

\modelentry{OCRNet}{ocrnet}
& \basescore{75.39}{0.79} & \scscore{77.88}{0.43}{2.49}
& \basescore{82.62}{0.11} & \scscore{83.03}{0.12}{0.41}
& \basescore{44.26}{0.66} & \scscore{44.84}{0.32}{0.58} \\

\modelentry{InternImage}{internimage}
& \basescore{81.95}{0.11} & \scscore{82.35}{0.05}{0.40}
& \basescore{86.22}{0.08} & \scscore{86.35}{0.00}{0.13}
& \basescore{51.87}{0.18} & \scscore{53.11}{0.20}{1.24} \\

\modelentry{DOCNet}{docnet}
& \basescore{80.80}{0.32} & \scscore{81.47}{0.18}{0.67}
& \basescore{85.61}{0.05} & \scscore{85.77}{0.06}{0.16}
& \basescore{50.92}{0.50} & \scscore{51.62}{0.03}{0.70} \\

\modelentry{Flash-InternImage}{flashinternimage}
& \basescore{82.38}{0.13} & \scscore{82.48}{0.09}{0.10}
& \basescore{86.37}{0.06} & \scscore{86.49}{0.01}{0.12}
& \basescore{54.04}{0.08} & \scscore{54.59}{0.06}{0.55} \\

\modelentry{TransNeXt}{transnext}
& \basescore{82.45}{0.45} & \scscore{82.96}{0.20}{0.51}
& \basescore{86.66}{0.10} & \scscore{86.85}{0.05}{0.19}
& \basescore{54.98}{0.17} & \scscore{55.77}{0.28}{0.79} \\

\modelentry{UniRepLKNet}{unireplknet}
& \basescore{81.71}{0.05} & \scscore{82.06}{0.07}{0.35}
& \basescore{86.33}{0.12} & \scscore{86.47}{0.03}{0.14}
& \basescore{52.04}{0.14} & \scscore{52.50}{0.22}{0.46} \\

\modelentry{LOGCAN++}{logcanpp}
& \basescore{80.97}{0.08} & \scscore{81.38}{0.08}{0.41}
& \basescore{86.07}{0.06} & \scscore{86.30}{0.06}{0.23}
& \basescore{50.59}{0.13} & \scscore{52.85}{0.13}{2.26} \\

\bottomrule
\end{tabular}%
}
\vspace{-0.1cm}
\label{tab3}
\end{table}

\begin{table}[!ht]
\caption{Comparison of \textbf{\textit{ShadowCast}} with additional shadow augmentation baselines using DeepLabV3+ on Vaihingen, Potsdam, and LoveDA.}
\vspace{-0.2cm}
\label{new:shadowaugseries}
\centering
\resizebox{0.68\columnwidth}{!}{
\setlength{\tabcolsep}{1.5pt}
\begin{tabular}{@{}lccc@{}}
\toprule
Augmentation & Vaihingen & Potsdam & LoveDA \\
\midrule
No shadow augmentation & 77.53$\pm$0.30 & 82.95$\pm$0.10 & 43.29$\pm$0.77 \\
RandomShadow {\tiny\cite{buslaev2020albumentations}}
 & 78.36$\pm$0.04 & 83.15$\pm$0.12 & 43.67$\pm$0.40 \\
 PlasmaShadow {\tiny\cite{buslaev2020albumentations}}
 & 76.98$\pm$0.19 & 82.37$\pm$0.11 & 44.01$\pm$0.17 \\
 SynShadow {\tiny\cite{SynShadow_inoue2020learning}}
 & 78.18$\pm$0.05 & 83.11$\pm$0.14 & 43.58$\pm$0.33 \\
SDAA {\tiny\cite{SDAA_mohajerani2021cloud}}
 & 77.87$\pm$0.08 & 83.21$\pm$0.12 & 43.12$\pm$0.60 \\
Zhong et al. {\tiny\cite{Zhong_zhong2022shadows}}
 & 78.07$\pm$0.19 & 83.24$\pm$0.12 & 43.46$\pm$0.56 \\
SatelliteCloudGenerator \tiny{\cite{Cloud_czerkawski2023satellitecloudgenerator}}
 & 77.44$\pm$0.28 & 82.87$\pm$0.21 & 43.28$\pm$0.56 \\
\midrule\rowcolor[HTML]{FFCE93}
\textbf{\textit{ShadowCast}}
 & \textbf{78.46$\pm$0.19} & \textbf{83.32$\pm$0.09} & \textbf{44.89$\pm$0.09} \\
\bottomrule
\end{tabular}
}
\vspace{-0.2cm}
\end{table}

\vspace{-0.3cm}
\section{Experiments}

\vspace{-0.1cm}
\subsection{Experimental Setup}

We evaluate \textbf{\textit{ShadowCast}} on three remote sensing semantic segmentation benchmarks, ISPRS Vaihingen, ISPRS Potsdam \cite{vaihingen}, and LoveDA \cite{wang2021loveda}. 
    Mean intersection-over-union (mIoU) is used as the evaluation metric, excluding the clutter class for Vaihingen and Potsdam.
    Unless otherwise specified, the results are reported as mean $\pm$ standard deviation over three independent runs.
For constructing the elevated-object mask $\mathbf{M}$ in \textbf{\textit{ShadowCast}}, $\mathcal{E}=\{\text{Building},\text{Tree}\}$ for ISPRS Vaihingen and Potsdam, while $\mathcal{E}=\{\text{Building},\text{Forest}\}$ for LoveDA.

\vspace{-0.3cm}
\subsection{Ablation Study}

\subsubsection{Parameter Sensitivity} 
A detailed sensitivity analysis of the main \textbf{\textit{ShadowCast}} hyperparameters on all benchmarks using DeepLabV3+ is provided in Tables~S1-S3 of the \textcolor{gray}{Supplementary Material}.
    The results show stable performance across the tested settings and identify $(d_{\min},d_{\max})=(5,30)$, $(\delta_{\min},\delta_{\max})=(0.5,0.8)$, $\eta=15$, $\tau=50$, and $r=8$ as the best-performing configuration, on all datasets.
Accordingly, we use these values together with $p=0.5$ as a single default configuration for all subsequent experiments.

\subsubsection{Effect of Cast-Shadow Direction Selection} Table~\ref{tab2} investigates the effect of image-specific cast-shadow direction estimation. 
While both random and estimated directions improve over the baseline, combining the estimated direction with angular jittering consistently achieves the best performance on all three datasets. 
    This indicates that image-specific direction estimation provides useful cues, while angular perturbation introduces extra diversity to boost model robustness.

\vspace{-0.3cm}
\subsection{Comparison across Segmentation Architectures}

Table~\ref{tab3} evaluates \textbf{\textit{ShadowCast}} across ten semantic segmentation architectures on all three benchmarks. 
    Across the evaluated model-dataset combinations, \textbf{\textit{ShadowCast}} generally yields improved performance over the corresponding baselines, with more pronounced gains observed on Vaihingen and LoveDA. 
Notably, OCRNet on Vaihingen and LOGCAN++ on LoveDA improve by 2.49 and 2.26 mIoU, respectively.

\vspace{-0.3cm}
\subsection{Comparison with Existing Shadow Augmentation Methods}
To further evaluate the effectiveness of \textbf{\textit{ShadowCast}}, we compare it with six existing shadow augmentation methods using DeepLab V3+ under the same training strategy; detailed settings are in the \textcolor{gray}{Supplementary Material}.
    As indicated in Table~\ref{new:shadowaugseries}, \textbf{\textit{ShadowCast}} achieves the highest mIoU across all three benchmarks.
        In contrast, existing shadow augmentation methods show less consistent improvements.
Overall, these results suggest incorporating semantic and photometric context into shadow augmentation, rather than generating shadows independently of scene content.

\begin{table}[t!]
\vspace{-0.1cm}
\centering
\caption{Impact of combining \textit{\textbf{ShadowCast}} with existing
augmentation/regularization methods on DeepLabV3+ for Vaihingen, Potsdam, and
LoveDA.} \vspace{-0.2cm}

\newcommand{\augplain}[1]{%
\shortstack[l]{#1\\{\phantom{[00]}}}%
}
\newcommand{\augmethod}[2]{%
\shortstack[l]{#1 {\tiny\cite{#2}} \\{\phantom{[00]}}}%
}
\newcommand{\augbase}[2]{%
\shortstack[c]{#1$\pm$#2\\{\phantom{(+0.00)}}}%
}
\newcommand{\augscore}[3]{%
\cellcolor[HTML]{FFCE93}%
\shortstack[c]{\textbf{#1$\pm$#2}\\{\textbf{(\textcolor{red}{+#3})}}}%
}

\resizebox{0.97\columnwidth}{!}{%
\setlength{\tabcolsep}{1.5pt}
\renewcommand{\arraystretch}{1.05}
\begin{tabular}{lcc cc cc}
\toprule
\multirow{2}{*}{Existing method}
& \multicolumn{2}{c}{Vaihingen}
& \multicolumn{2}{c}{Potsdam}
& \multicolumn{2}{c}{LoveDA} \\
\cmidrule(lr){2-3}\cmidrule(lr){4-5}\cmidrule(lr){6-7}
& Baseline & \cellcolor[HTML]{FFCE93}\textbf{\textit{ShadowCast}}
& Baseline & \cellcolor[HTML]{FFCE93}\textbf{\textit{ShadowCast}}
& Baseline & \cellcolor[HTML]{FFCE93}\textbf{\textit{ShadowCast}} \\
\midrule

\augplain{Horizontal Flip}
& \augbase{78.62}{0.19} & \augscore{79.08}{0.06}{0.46}
& \augbase{84.00}{0.12} & \augscore{84.32}{0.08}{0.32}
& \augbase{43.81}{0.60} & \augscore{45.88}{0.52}{2.07} \\

\augplain{Random Resize}
& \augbase{80.04}{0.38} & \augscore{80.36}{0.02}{0.32}
& \augbase{84.37}{0.13} & \augscore{84.50}{0.03}{0.13}
& \augbase{45.11}{0.31} & \augscore{46.70}{0.97}{1.59} \\

\augplain{Photometric Distortion}
& \augbase{77.33}{0.35} & \augscore{77.86}{0.12}{0.53}
& \augbase{82.76}{0.13} & \augscore{83.13}{0.05}{0.37}
& \augbase{47.19}{0.64} & \augscore{48.45}{0.21}{1.26} \\

\augmethod{CLAHE}{zuiderveld1994contrast}
& \augbase{78.33}{0.12} & \augscore{78.46}{0.04}{0.13}
& \augbase{83.13}{0.25} & \augscore{83.30}{0.27}{0.17}
& \augbase{43.92}{0.78} & \augscore{44.59}{0.49}{0.67} \\

\augmethod{CutOut}{devries2017improved}
& \augbase{78.29}{0.69} & \augscore{79.18}{0.30}{0.89}
& \augbase{83.27}{0.03} & \augscore{83.56}{0.07}{0.29}
& \augbase{44.54}{0.46} & \augscore{46.27}{0.75}{1.73} \\

\augmethod{CutMix}{yun2019cutmix}
& \augbase{77.54}{0.60} & \augscore{78.39}{0.10}{0.85}
& \augbase{82.99}{0.05} & \augscore{83.35}{0.03}{0.36}
& \augbase{45.54}{0.66} & \augscore{46.30}{0.74}{0.76} \\

\augmethod{Random Erasing}{zhong2020random}
& \augbase{79.01}{0.27} & \augscore{79.50}{0.20}{0.49}
& \augbase{83.43}{0.04} & \augscore{83.70}{0.05}{0.27}
& \augbase{43.78}{1.42} & \augscore{45.82}{0.32}{2.04} \\

\augmethod{Label Smoothing}{szegedy2016rethinking}
& \augbase{77.97}{0.28} & \augscore{78.66}{0.21}{0.69}
& \augbase{82.84}{0.13} & \augscore{83.33}{0.06}{0.49}
& \augbase{43.82}{1.29} & \augscore{46.46}{0.23}{2.64} \\

\augmethod{NSegment}{kim2025nsegment}
& \augbase{78.26}{0.63} & \augscore{78.62}{0.21}{0.36}
& \augbase{83.16}{0.11} & \augscore{83.37}{0.12}{0.21}
& \augbase{43.36}{0.62} & \augscore{45.56}{0.61}{2.20} \\

\bottomrule
\end{tabular}%
}
\vspace{-0.1cm}
\label{tab4}
\end{table}


\begin{table}[!t]
\centering
\caption{
Evaluation of \textbf{\textit{ShadowCast}} using DeepLab V3+ across shadowed and non-shadowed regions of Vaihingen, Potsdam, and LoveDA (Ten shadow-annotated test images per dataset).
} \vspace{-0.2cm}
\label{new:shadow}
\resizebox{0.68\columnwidth}{!}{
\setlength{\tabcolsep}{3pt}
\begin{tabular}{@{}llll@{}}
\toprule
Dataset & Method & {Shadowed} & {Non-shadowed} \\
\midrule
Vaihingen & Baseline & 57.88$\pm$0.95 & 82.77$\pm$0.37 \\
 & \cellcolor[HTML]{FFCE93}\textbf{\textit{ShadowCast}}
 & \cellcolor[HTML]{FFCE93}\textbf{58.18$\pm$0.48 (\textcolor{red}{+0.30})}
 & \cellcolor[HTML]{FFCE93}\textbf{83.26$\pm$0.33 (\textcolor{red}{+0.49})} \\
\midrule
Potsdam & Baseline & 81.85$\pm$0.93 & 81.50$\pm$3.00 \\
 & \cellcolor[HTML]{FFCE93}\textbf{\textit{ShadowCast}}
 & \cellcolor[HTML]{FFCE93}\textbf{82.27$\pm$0.13 (\textcolor{red}{+0.42})}
 & \cellcolor[HTML]{FFCE93}\textbf{82.85$\pm$2.07 (\textcolor{red}{+1.35})} \\
\midrule
LoveDA & Baseline & 26.83$\pm$1.34 & 38.86$\pm$1.48 \\
 & \cellcolor[HTML]{FFCE93}\textbf{\textit{ShadowCast}}
 & \cellcolor[HTML]{FFCE93}\textbf{27.14$\pm$0.71 (\textcolor{red}{+0.31})}
 & \cellcolor[HTML]{FFCE93}\textbf{39.41$\pm$1.13 (\textcolor{red}{+0.55})} \\
\bottomrule
\end{tabular}
} \vspace{-0.2cm}
\end{table}

\vspace{-0.35cm}
\subsection{Compatibility with Existing Training Strategies}

Table~\ref{tab4} evaluates \textbf{\textit{ShadowCast}} in combination with existing augmentation and regularization strategies using DeepLabV3+. 
    Incorporating \textbf{\textit{ShadowCast}} improves performance across all tested configurations and datasets. 
This improvement is observed when combined with conventional geometric and photometric augmentations, indicating that the proposed shadow synthesis provides additional variation beyond standard image transformations. 
    \textbf{\textit{ShadowCast}} is also compatible with regularization-oriented strategies, including label smoothing \cite{szegedy2016rethinking}, which replaces hard one-hot targets with softened class distributions to mitigate overconfident predictions, and NSegment \cite{kim2025nsegment}, which applies controlled elastic deformations only to segmentation labels to improve robustness against annotation inconsistencies. 
These results demonstrate that \textbf{\textit{ShadowCast}} provides a complementary training signals that can be combined with other training tricks.

\vspace{-0.35cm}
\subsection{Effects across Shadow and Non-Shadow Regions}

To assess the effectiveness of \textbf{\textit{ShadowCast}} under real cast shadows, we separately evaluate segmentation performance in shadowed and non-shadowed regions.
    Since none of three benchmarks provides shadow annotations, we manually annotated pixel-level shadow masks for ten images chosen from each dataset; further details are in the \textcolor{gray}{Supplementary Material}.
As shown in Table~\ref{new:shadow}, \textbf{\textit{ShadowCast}} improves mIoU in shadowed regions by 0.30, 0.42, and 0.31 on Vaihingen, Potsdam, and LoveDA, respectively, while achieving larger gains of 0.49, 1.35, and 0.55 points in non-shadowed regions.
    These findings suggest that synthetic shadow augmentation can improve robustness to real shadows without sacrificing segmentation performance in non-shadowed regions.
Although the labor-intensive nature of pixel-level shadow annotation limited our evaluation to ten images per dataset, consistent improvements observed in both regions support the effectiveness of our method.

\vspace{-0.cm}
\section{Conclusions}

We proposed \textbf{\textit{ShadowCast}}, a lightweight and learning-free augmentation method for improving the robustness of remote sensing semantic segmentation to cast-shadow variations. 
    By leveraging semantic labels and an image-specific cast-shadow direction cue, \textbf{\textit{ShadowCast}} generates context-aware synthetic shadows without requiring additional learned models, shadow annotations, or acquisition metadata. 
Experiments across multiple datasets and segmentation architectures demonstrate consistent improvements over standard training and existing shadow augmentation methods.
    Furthermore, evaluations on shadowed and non-shadowed regions show that its benefits extend beyond regions directly affected by cast shadows.

\noindent \textbf{Limitations.} Despite these advantages, the proposed method relies on brightness-based direction estimation and simplified mask translation, which may produce inaccurate shadows in complex scenes or terrains with significant elevation variations, as discussed in \textcolor{gray}{Supplementary Material}. 
    Nevertheless, the random-direction ablation in Table~\ref{tab2} shows improvements over the baseline across all three benchmarks, suggesting that occasional direction estimation errors may not necessarily negate the benefits of shadow augmentation.




\ifCLASSOPTIONcaptionsoff
  \newpage
\fi

\balance


%








\newpage

\onecolumn

\setcounter{section}{0}
\setcounter{table}{0}
\setcounter{figure}{0}

\renewcommand{\thetable}{S\arabic{table}}
\renewcommand{\thefigure}{S\arabic{figure}}

\newcommand{\dft}[1]{\textcolor{gray!55}{#1}}

{\Huge\textbf{Appendix}}

\section{Parameter Sensitivity Analysis}

We conduct a hyperparameter sensitivity analysis of \textbf{\textit{ShadowCast}} on ISPRS Vaihingen, ISPRS Potsdam, and LoveDA with DeepLabV3+, with the transform probability fixed at (p=0.5).
Tables~\ref{tab:sensitivity},
\ref{tab:sensitivity_potsdam}, and \ref{tab:sensitivity_loveda} report the
mean intersection-over-union (mIoU) as mean\,$\pm$\,standard deviation over
three independent runs. 
    In each row, \textcolor{gray}{gray}-color entries indicate parameters held at their default values, whereas black-color entries indicate the parameter being varied.

\begin{table}[H]
\centering
\caption{Parameter sensitivity of \textbf{\textit{ShadowCast}} on the ISPRS
Vaihingen benchmark using DeepLabV3+ ($p=0.5$).}
\label{tab:sensitivity}
\setlength{\tabcolsep}{9pt}
\renewcommand{\arraystretch}{1.15}

\begin{tabular}{cccccc}
\toprule
$(d_{\min},d_{\max})$ & $(\delta_{\min},\delta_{\max})$ & $\eta$ & $\tau$ & $r$ & mIoU \\
\midrule
\textbf{(5, 30)} & \textbf{(0.5, 0.8)} & \textbf{15} & \textbf{50} & \textbf{8} & \textbf{78.46\,$\pm$\,0.19} \\
\midrule
(5, 20) & \dft{(0.5, 0.8)} & \dft{15} & \dft{50} & \dft{8} & 78.25\,$\pm$\,0.33 \\
(5, 40) & \dft{(0.5, 0.8)} & \dft{15} & \dft{50} & \dft{8} & 78.22\,$\pm$\,0.18 \\
\midrule
\dft{(5, 30)} & (0.4, 0.7) & \dft{15} & \dft{50} & \dft{8} & 78.35\,$\pm$\,0.12 \\
\dft{(5, 30)} & (0.6, 0.9) & \dft{15} & \dft{50} & \dft{8} & 78.10\,$\pm$\,0.36 \\
\midrule
\dft{(5, 30)} & \dft{(0.5, 0.8)} & 0  & \dft{50} & \dft{8} & 78.20\,$\pm$\,0.07 \\
\dft{(5, 30)} & \dft{(0.5, 0.8)} & 5  & \dft{50} & \dft{8} & 78.34\,$\pm$\,0.37 \\
\dft{(5, 30)} & \dft{(0.5, 0.8)} & 10 & \dft{50} & \dft{8} & 78.41\,$\pm$\,0.18 \\
\dft{(5, 30)} & \dft{(0.5, 0.8)} & 20 & \dft{50} & \dft{8} & 78.34\,$\pm$\,0.06 \\
\midrule
\dft{(5, 30)} & \dft{(0.5, 0.8)} & \dft{15} & 0  & \dft{8} & 78.15\,$\pm$\,0.33 \\
\dft{(5, 30)} & \dft{(0.5, 0.8)} & \dft{15} & 25 & \dft{8} & 78.29\,$\pm$\,0.27 \\
\dft{(5, 30)} & \dft{(0.5, 0.8)} & \dft{15} & 75 & \dft{8} & 78.23\,$\pm$\,0.10 \\
\midrule
\dft{(5, 30)} & \dft{(0.5, 0.8)} & \dft{15} & \dft{50} & 6  & 78.17\,$\pm$\,0.27 \\
\dft{(5, 30)} & \dft{(0.5, 0.8)} & \dft{15} & \dft{50} & 10 & 78.04\,$\pm$\,0.14 \\
\bottomrule
\end{tabular}
\end{table}

\begin{table}[H]
\centering
\caption{Parameter sensitivity of \textbf{\textit{ShadowCast}} on the ISPRS
Potsdam benchmark using DeepLabV3+ ($p=0.5$).}
\label{tab:sensitivity_potsdam}
\setlength{\tabcolsep}{9pt}
\renewcommand{\arraystretch}{1.15}
\begin{tabular}{cccccc}
\toprule
$(d_{\min},d_{\max})$ & $(\delta_{\min},\delta_{\max})$ & $\eta$ & $\tau$ & $r$ & mIoU \\
\midrule
\textbf{(5, 30)} & \textbf{(0.5, 0.8)} & \textbf{15} & \textbf{50} & \textbf{8} & \textbf{83.32\,$\pm$\,0.09} \\
\midrule
(5, 20) & \dft{(0.5, 0.8)} & \dft{15} & \dft{50} & \dft{8} & 83.24\,$\pm$\,0.13 \\
(5, 40) & \dft{(0.5, 0.8)} & \dft{15} & \dft{50} & \dft{8} & 83.27\,$\pm$\,0.11 \\
\midrule
\dft{(5, 30)} & (0.4, 0.7) & \dft{15} & \dft{50} & \dft{8} & 83.26\,$\pm$\,0.12 \\
\dft{(5, 30)} & (0.6, 0.9) & \dft{15} & \dft{50} & \dft{8} & 83.18\,$\pm$\,0.11 \\
\midrule
\dft{(5, 30)} & \dft{(0.5, 0.8)} & 0  & \dft{50} & \dft{8} & 83.22\,$\pm$\,0.13 \\
\dft{(5, 30)} & \dft{(0.5, 0.8)} & 5  & \dft{50} & \dft{8} & 83.21\,$\pm$\,0.20 \\
\dft{(5, 30)} & \dft{(0.5, 0.8)} & 10 & \dft{50} & \dft{8} & 83.27\,$\pm$\,0.16 \\
\dft{(5, 30)} & \dft{(0.5, 0.8)} & 20 & \dft{50} & \dft{8} & 83.28\,$\pm$\,0.22 \\
\midrule
\dft{(5, 30)} & \dft{(0.5, 0.8)} & \dft{15} & 0  & \dft{8} & 83.21\,$\pm$\,0.10 \\
\dft{(5, 30)} & \dft{(0.5, 0.8)} & \dft{15} & 25 & \dft{8} & 83.30\,$\pm$\,0.04 \\
\dft{(5, 30)} & \dft{(0.5, 0.8)} & \dft{15} & 75 & \dft{8} & 83.20\,$\pm$\,0.21 \\
\midrule
\dft{(5, 30)} & \dft{(0.5, 0.8)} & \dft{15} & \dft{50} & 6  & 83.22\,$\pm$\,0.10 \\
\dft{(5, 30)} & \dft{(0.5, 0.8)} & \dft{15} & \dft{50} & 10 & 83.23\,$\pm$\,0.16 \\
\bottomrule
\end{tabular}
\end{table}

\begin{table}[H]
\centering
\caption{Parameter sensitivity of \textbf{\textit{ShadowCast}} on the LoveDA
benchmark using DeepLabV3+ ($p=0.5$).}
\label{tab:sensitivity_loveda}
\setlength{\tabcolsep}{9pt}
\renewcommand{\arraystretch}{1.15}
\begin{tabular}{cccccc}
\toprule
$(d_{\min},d_{\max})$ & $(\delta_{\min},\delta_{\max})$ & $\eta$ & $\tau$ & $r$ & mIoU \\
\midrule
\textbf{(5, 30)} & \textbf{(0.5, 0.8)} & \textbf{15} & \textbf{50} & \textbf{8} & \textbf{44.89\,$\pm$\,0.09} \\
\midrule
(5, 20) & \dft{(0.5, 0.8)} & \dft{15} & \dft{50} & \dft{8} & 44.15\,$\pm$\,0.23 \\
(5, 40) & \dft{(0.5, 0.8)} & \dft{15} & \dft{50} & \dft{8} & 44.20\,$\pm$\,0.21 \\
\midrule
\dft{(5, 30)} & (0.4, 0.7) & \dft{15} & \dft{50} & \dft{8} & 44.29\,$\pm$\,0.47 \\
\dft{(5, 30)} & (0.6, 0.9) & \dft{15} & \dft{50} & \dft{8} & 44.21\,$\pm$\,0.23 \\
\midrule
\dft{(5, 30)} & \dft{(0.5, 0.8)} & 0  & \dft{50} & \dft{8} & 44.45\,$\pm$\,0.30 \\
\dft{(5, 30)} & \dft{(0.5, 0.8)} & 5  & \dft{50} & \dft{8} & 44.49\,$\pm$\,0.47 \\
\dft{(5, 30)} & \dft{(0.5, 0.8)} & 10 & \dft{50} & \dft{8} & 44.58\,$\pm$\,0.42 \\
\dft{(5, 30)} & \dft{(0.5, 0.8)} & 20 & \dft{50} & \dft{8} & 44.47\,$\pm$\,0.26 \\
\midrule
\dft{(5, 30)} & \dft{(0.5, 0.8)} & \dft{15} & 0  & \dft{8} & 43.70\,$\pm$\,0.43 \\
\dft{(5, 30)} & \dft{(0.5, 0.8)} & \dft{15} & 25 & \dft{8} & 44.20\,$\pm$\,0.15 \\
\dft{(5, 30)} & \dft{(0.5, 0.8)} & \dft{15} & 75 & \dft{8} & 44.39\,$\pm$\,0.18 \\
\midrule
\dft{(5, 30)} & \dft{(0.5, 0.8)} & \dft{15} & \dft{50} & 6  & 44.31\,$\pm$\,0.51 \\
\dft{(5, 30)} & \dft{(0.5, 0.8)} & \dft{15} & \dft{50} & 10 & 44.23\,$\pm$\,0.19 \\
\bottomrule
\end{tabular}
\end{table}

The results remain stable across the evaluated parameter ranges on all three datasets, with all variants on Potsdam staying within 0.14 mIoU of the default. 
    LoveDA, which covers more diverse urban and rural scenes, shows a somewhat higher sensitivity: all variants stay within 0.74 mIoU of the default except for $\tau=0$, which causes the largest drop (1.19 mIoU).
This indicates that the region-based context check becomes more important as scene diversity increases. 
    Moderate shadow displacement and attenuation settings generally perform better, while angular jittering improves robustness relative to using no perturbation. 
On all three benchmarks, the best performance (78.46 mIoU on Vaihingen, 83.32 mIoU on Potsdam, and 44.89 mIoU on LoveDA, in average) is obtained with $(d_{\min},d_{\max})=(5,30)$, $(\delta_{\min},\delta_{\max})=(0.5,0.8)$, $\eta=15$, $\tau=50$, and $r=8$.
    We therefore use this setting as the default configuration in all experiments reported in the remaining experiments.
These results also indicate that a single configuration of the pixel-defined parameters is appropriate for all three datasets despite their different spatial resolutions.

\section{Detailed Settings for Evaluation on Shadowed and Non-Shadowed Regions}

In Section IV-F of the main manuscript, to
examine how \textbf{\textit{ShadowCast}} affects segmentation with respect to real cast shadows, we evaluate shadowed and non-shadowed regions separately.
    Unfortunately, none of the three benchmarks provides shadow annotations.
Since pixel-level shadow annotation is labor-intensive, we selected 10 test images per dataset that contain cast shadows and manually annotated their shadow regions at the pixel level; see Fig.~\ref{supp:fig_shadow} for instance. 
    The resulting masks divide the pixels of these images into a shadowed and a non-shadowed region. 
For each region, the class-wise IoU is computed using only the pixels within that region and averaged over the evaluated classes (clutter excluded for
Vaihingen and Potsdam). 
    For each dataset, we evaluate previously trained DeepLabV3+ checkpoints from three independent runs for both the baseline and \textbf{\textit{ShadowCast}}. 
We then report mIoU separately for shadowed and non-shadowed regions on the ten chosen and shadow-annotated images per each dataset.

\providecommand{\suppimg}[2]{%
  \IfFileExists{#2}{%
    \ifdefined\includegraphics
      \includegraphics[width=#1]{#2}%
    \else
      \pdfximage width #1 {#2}\pdfrefximage\pdflastximage
    \fi}%
  {\fbox{\parbox[c][#1][c]{#1}{\centering\scriptsize missing: \detokenize{#2}}}}}
  
\begin{figure}[!htb]
\centering
\setlength{\tabcolsep}{3pt}
\begin{tabular}{ccc}
\suppimg{0.30\linewidth}{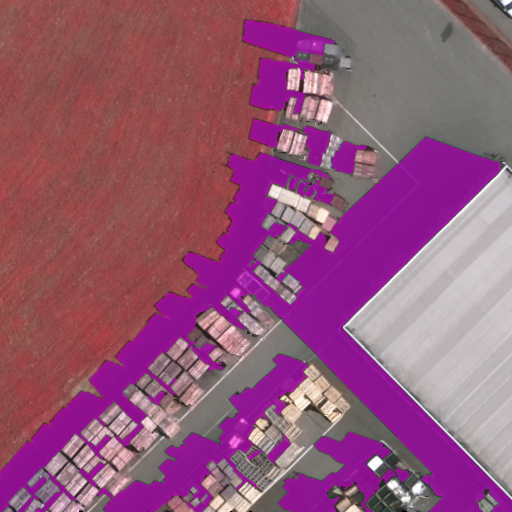} &
\suppimg{0.30\linewidth}{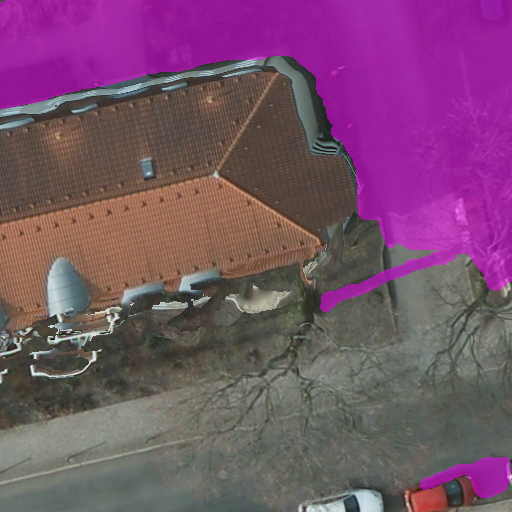} &
\suppimg{0.30\linewidth}{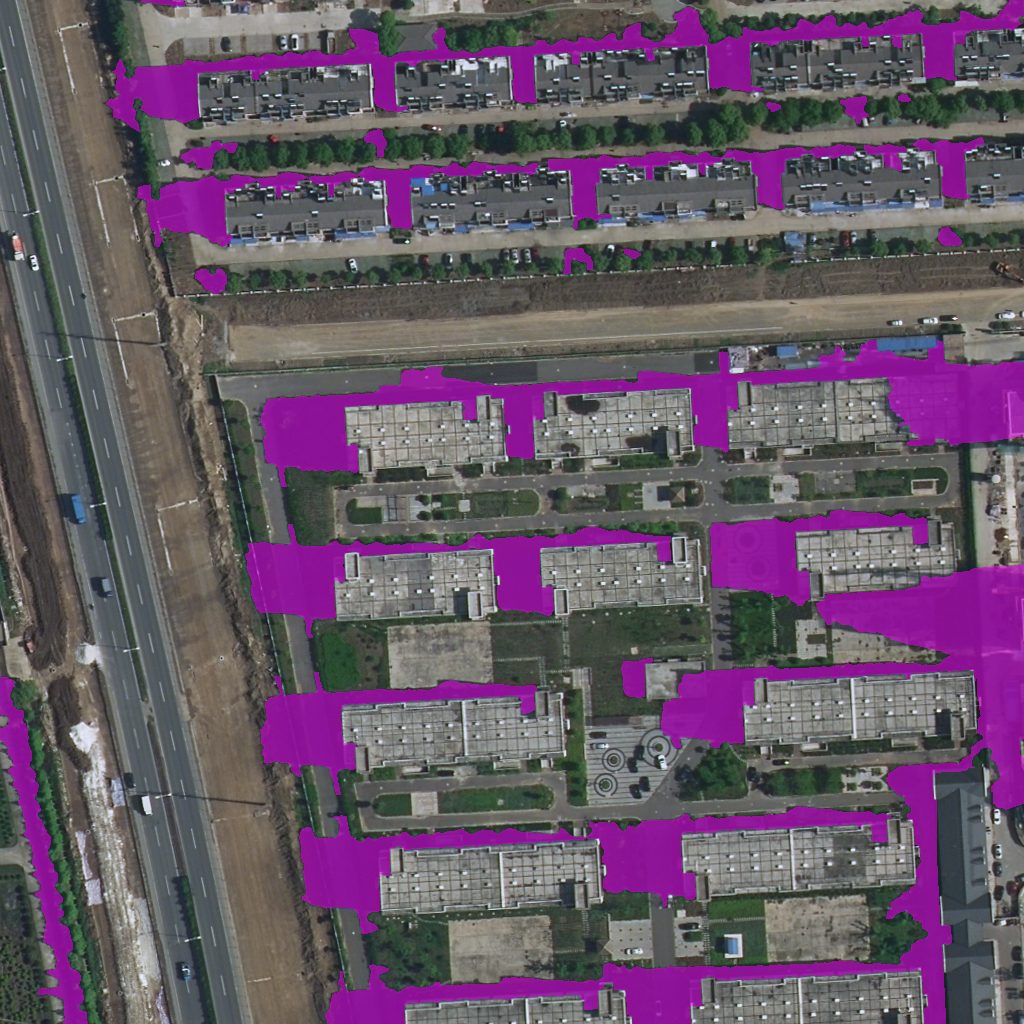} \\
{\footnotesize (a) Vaihingen} & {\footnotesize (b) Potsdam} & {\footnotesize (c) LoveDA}
\end{tabular}
\caption{Examples of the manually annotated real shadow regions (\textcolor{magenta}{magenta}-colored) on
(a) Vaihingen, (b) Potsdam, and (c) LoveDA.}
\label{supp:fig_shadow}
\end{figure}

\section{Comparison with Existing Shadow Augmentation Methods}
\label{sec:supp_baselines}

This section provides qualitative examples of the existing shadow augmentation methods evaluated in Table~III of the main manuscript, along with their implementation details and parameter settings.

\begin{figure}[!htb]
\centering
\setlength{\tabcolsep}{2pt}
\begin{tabular}{cccc}
\includegraphics[width=0.235\linewidth]{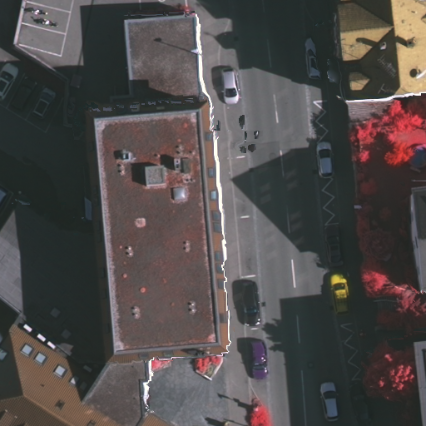} &
\includegraphics[width=0.235\linewidth]{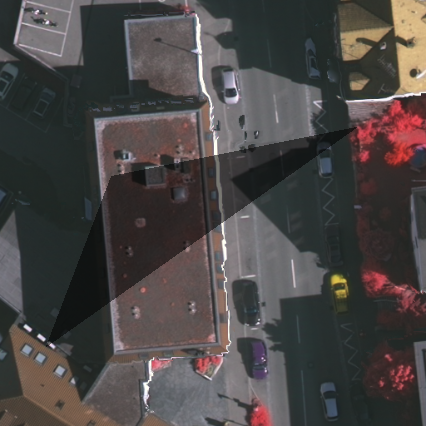} &
\includegraphics[width=0.235\linewidth]{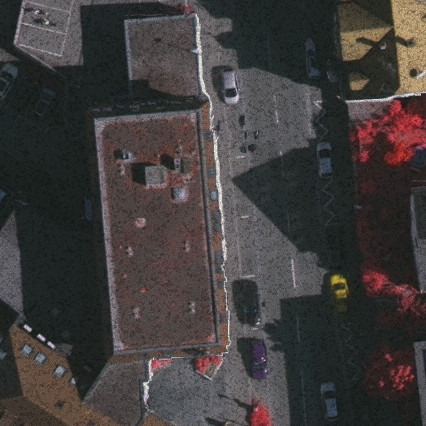} &
\includegraphics[width=0.235\linewidth]{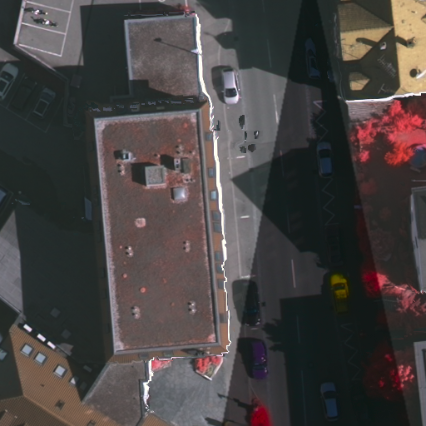} \\
{\footnotesize (a) Original} &
{\footnotesize (b) RandomShadow~\cite{buslaev2020albumentations}} &
{\footnotesize (c) PlasmaShadow~\cite{buslaev2020albumentations}} &
{\footnotesize (d) Zhong \textit{et al.}~\cite{Zhong_zhong2022shadows}} \\[4pt]
\end{tabular}
\begin{tabular}{cccc}
\includegraphics[width=0.235\linewidth]{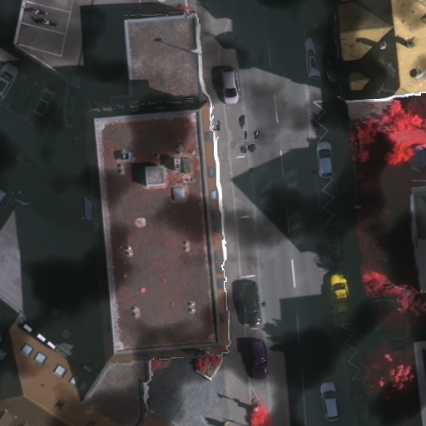} &
\includegraphics[width=0.235\linewidth]{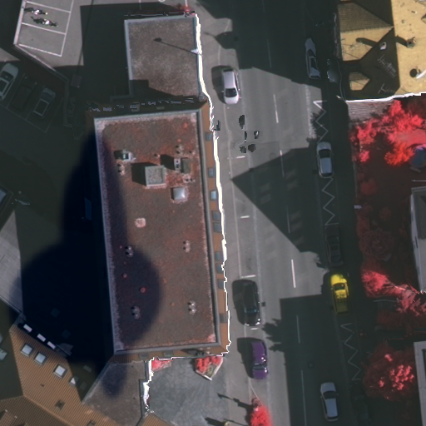} &
\includegraphics[width=0.235\linewidth]{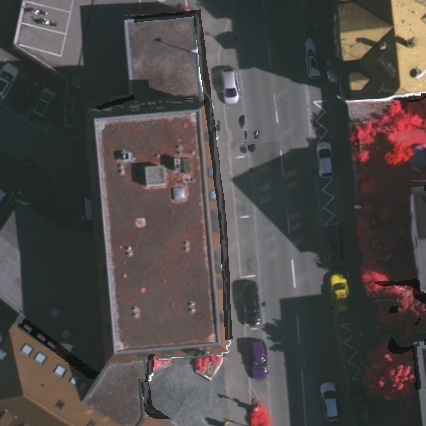} &
\includegraphics[width=0.235\linewidth]{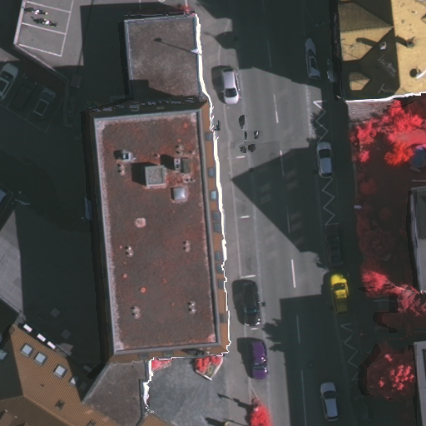} \\
{\footnotesize (e) SatelliteCloudGenerator~\cite{Cloud_czerkawski2023satellitecloudgenerator}} &
{\footnotesize (f) SynShadow~\cite{SynShadow_inoue2020learning}} &
{\footnotesize (g) SDAA~\cite{SDAA_mohajerani2021cloud}} &
{\footnotesize (h) \textbf{\textit{ShadowCast}} (Ours)} \\
\end{tabular}
\caption{Shadows synthesized by different augmentation methods on the same Vaihingen training image.
CLAHE~\cite{zuiderveld1994contrast}, which enhances the image rather than synthesizing shadows, is omitted.}
\label{fig:supp_cmp}
\end{figure}
 
\subsection{Compared Methods}
We group the compared methods by the scene context they use for shadow synthesis as follows.

\subsubsection{Object-Agnostic Shadow Synthesis}
RandomShadow~\cite{buslaev2020albumentations} applies uniform darkening to randomly generated polygonal regions. 
    PlasmaShadow~\cite{buslaev2020albumentations} attenuates the entire image using a plasma fractal pattern generated by the diamond-square algorithm.
Zhong \textit{et al.}~\cite{Zhong_zhong2022shadows} generate triangular shadows by scaling the lightness channel in the CIELAB color space. Their original method optimizes the triangle vertices using particle swarm optimization to construct adversarial examples; we instead sample the vertices randomly.
SatelliteCloudGenerator~\cite{Cloud_czerkawski2023satellitecloudgenerator} synthesizes cloud and cloud-shadow masks from multi-scale Perlin noise and models shadows as uncorrelated with the clouds visible in the image; we use only its shadow component.
SynShadow~\cite{SynShadow_inoue2020learning} composites shadow mattes rendered from 3D object models onto shadow-free images using a physically grounded shadow illumination model. The mattes, however, are placed independently of the target scene content.
 
\subsubsection{Object-Anchored Shadow Synthesis}
Sunlight Direction-Aware Augmentation (SDAA)~\cite{SDAA_mohajerani2021cloud} first removes existing shadows using shadow annotations, then synthesizes cloud shadows by translating cloud masks according to the solar azimuth and zenith angles provided in the image metadata. 
    To vary the shadow direction, it applies additional azimuth offsets of $\{90^\circ,180^\circ,270^\circ\}$.
Since neither solar angles nor shadow annotations are available for the three benchmarks, we adapt SDAA by translating the label-derived elevated-object mask $\mathbf{M}$, rather than cloud masks, along a direction sampled uniformly from $[0^\circ,360^\circ)$ for each image. 
    The adapted SDAA and \textbf{\textit{ShadowCast}} thus use the same shadow-casting objects, allowing a comparison of random and image-specific cast-shadow directions. 
Their darkening models also differ, however: SDAA uses a gamma transform, whereas \textbf{\textit{ShadowCast}} applies linear attenuation.

Note that their implementation details and parameter settings are given in Section~\ref{sec:supp_impl} of Supplementary Material.

 
 
\subsection{Qualitative Comparison}


Fig.~\ref{fig:supp_cmp} compares the synthetic shadows generated by each method on the same Vaihingen training image, where real cast shadows extend to the left of buildings and trees (Fig.~\ref{fig:supp_cmp}(a)). 
    The object-agnostic methods introduce dark regions without accounting for scene structure.
RandomShadow and Zhong \textit{et al.} darken polygonal and triangular regions whose straight boundaries cut across the rooftop, sunlit road, vehicles, and tree canopies, regardless of the locations of shadow-casting objects (Fig.~\ref{fig:supp_cmp}(b) and (d)). 
    PlasmaShadow applies a mottled fractal pattern across the entire image, reducing overall brightness and contrast rather than forming cast shadows adjacent to elevated objects (Fig.~\ref{fig:supp_cmp}(c)). 
SatelliteCloudGenerator darkens roughly half of the image with irregular patterns unrelated to these objects (Fig.~\ref{fig:supp_cmp}(e)).
        SynShadow produces a soft-edged shadow with physically modeled attenuation, but its shadow matte partially covers the rooftop: this would require a taller occluder on the building's sunward side, where none is visible (Fig.~\ref{fig:supp_cmp}(f)). 
SDAA uses semantic labels to anchor shadows to buildings and trees, but samples their direction at random. 
    In this example, it projects shadows toward the upper right, onto the sunlit side of the buildings and opposite to the real shadows (Fig.~\ref{fig:supp_cmp}(g)). 
Without shadow annotations to remove the existing shadows, the augmented image thus contains cast shadows in conflicting directions.
In contrast, \textbf{\textit{ShadowCast}} projects shadows from the same objects toward the left, matching the observed shadow direction (Fig.~\ref{fig:supp_cmp}(h)): synthesized appearance variations therefore remain consistent with the scene geometry and illumination.


\section{Implementation Details and Parameter Settings}
\label{sec:supp_impl}
 
This section reports the training configurations of the segmentation models
and the parameter settings of all augmentation and regularization methods
evaluated in the main manuscript and in this Supplementary Material. All
experiments are conducted with MMSegmentation (MMSeg) \footnote{\url{https://github.com/open-mmlab/mmsegmentation}}.
CLAHE, RandomShadow, and PlasmaShadow are taken from
Albumentations~\cite{buslaev2020albumentations}; horizontal flip, random
resize, photometric distortion, and CutOut are taken from MMSeg
(\texttt{RandomFlip}, \texttt{RandomResize}, \texttt{PhotoMetricDistortion},
and \texttt{RandomCutOut}); the remaining methods are implemented as MMSeg
transforms or, for label smoothing, as a loss function.
 
\subsection{Training Configuration}
 
Table~\ref{tab:supp_models} lists the backbone and optimizer of each
segmentation model in Table~II of the main manuscript. All models are trained
for 80k iterations, and the learning rate decays polynomially with power 0.9
to the minimum value in the table. SGD uses a momentum of 0.9, and AdamW uses
$\beta_1=0.9$. The baseline is trained without any of the methods in
Tables~\ref{tab:supp_aug} and~\ref{tab:supp_shadow}, and each of these methods
is added to the baseline training pipeline when it is evaluated.
 
\begin{table}[!htbp]
\centering
\caption{Training configurations of the segmentation models}
\label{tab:supp_models}
\begin{tabular}{@{}lllccc@{}}
\toprule
Model & Backbone & Optimizer & Initial LR & Weight decay & Minimum LR \\
\midrule
DeepLabV3+        & ResNet-18           & SGD   & 0.01      & $5\times10^{-4}$ & $10^{-4}$ \\
DANet             & ResNet-50           & SGD   & 0.01      & $5\times10^{-4}$ & $10^{-4}$ \\
APCNet            & ResNet-50           & SGD   & 0.01      & $5\times10^{-4}$ & $10^{-4}$ \\
OCRNet            & HRNet-W18-Small     & SGD   & 0.01      & $5\times10^{-4}$ & $10^{-4}$ \\
InternImage       & InternImage-T       & SGD   & 0.01      & $5\times10^{-4}$ & $10^{-4}$ \\
DOCNet            & HRNet-W32           & AdamW & $10^{-4}$ & $10^{-4}$        & 0 \\
Flash-InternImage & Flash-InternImage-T & SGD   & 0.01      & $5\times10^{-4}$ & $10^{-4}$ \\
TransNeXt         & TransNeXt-Tiny      & SGD   & 0.01      & $5\times10^{-4}$ & $10^{-4}$ \\
UniRepLKNet       & UniRepLKNet-T       & SGD   & 0.01      & $5\times10^{-4}$ & $10^{-4}$ \\
LOGCAN++          & RepViT-M2.3         & AdamW & $10^{-4}$ & $10^{-4}$        & 0 \\
\bottomrule
\end{tabular}
\end{table}
 
\subsection{Augmentation and Regularization Methods}
 
Table~\ref{tab:supp_aug} lists the settings of the methods combined with
\textbf{\textit{ShadowCast}} in Table~IV of the main manuscript. The custom
implementations are described below.
 
\begin{table}[!htbp]
\centering
\caption{Settings of the augmentation and regularization methods}
\label{tab:supp_aug}
\footnotesize
\begin{tabular}{@{}llp{0.6\linewidth}@{}}
\toprule
Method & Implementation & Parameters \\
\midrule
Horizontal Flip
  & MMSeg
  & \raggedright $p=0.5$ \tabularnewline
Random Resize
  & MMSeg
  & \raggedright Scale ratio $\sim\mathcal{U}(0.5,2.0)$ relative to $512\times512$
    ($1024\times1024$ for LoveDA); aspect ratio preserved \tabularnewline
Photometric Distortion
  & MMSeg
  & \raggedright Brightness shift $\sim\mathcal{U}(-32,32)$; contrast and saturation
    factors $\sim\mathcal{U}(0.5,1.5)$; integer hue shift $\in\{-18,\ldots,17\}$
    (OpenCV hue scale); each applied with probability 0.5 \tabularnewline
CLAHE~\cite{zuiderveld1994contrast}
  & Albumentations
  & \raggedright $p=0.5$; clip limit $\sim\mathcal{U}(1,4)$; $8\times8$ tiles; applied to
    the CIELAB lightness channel \tabularnewline
CutOut~\cite{devries2017improved}
  & MMSeg
  & \raggedright $p=0.5$; 5-10 holes of $16\times16$ or $32\times32$ pixels; image filled
    with 0 and labels with the ignore index 255 \tabularnewline
CutMix~\cite{yun2019cutmix}
  & Custom
  & \raggedright $p=0.5$; patch area ratio $\sim\mathcal{U}(0.2,0.5)$; image and label
    patches from another training sample \tabularnewline
Random Erasing~\cite{zhong2020random}
  & Custom
  & \raggedright $p=0.5$; area ratio $\sim\mathcal{U}(0.05,0.2)$; aspect ratio
    $\sim\mathcal{U}(0.5,2.0)$; filled with pixels resampled from the erased
    region; labels set to 255 \tabularnewline
Label Smoothing~\cite{szegedy2016rethinking}
  & Custom (Loss)
  & \raggedright $\epsilon=0.1$ \tabularnewline
NSegment~\cite{kim2025nsegment}
  & Custom
  & \raggedright $p=0.5$; $(\alpha,\sigma)\in\{1,15,30,50,100\}\times\{3,5,10\}$; label map
    only \tabularnewline
\bottomrule
\end{tabular}
\end{table}
 
\subsubsection{CutMix}
A rectangle with area ratio $\rho\sim\mathcal{U}(0.2,0.5)$, i.e., with side
lengths $\sqrt{\rho}\,H$ and $\sqrt{\rho}\,W$ for an $H\times W$ image, is
centered at a uniformly sampled pixel and clipped at the image border. The
image and label inside the rectangle are replaced by the same region of
another training sample.
 
\subsubsection{Random Erasing}
A rectangle with area ratio $\sim\mathcal{U}(0.05,0.2)$ and height-to-width
ratio $\sim\mathcal{U}(0.5,2.0)$ is placed uniformly inside the image, with up
to 10 attempts. Each channel inside the rectangle is filled with values drawn
with replacement from the same channel of the erased region, and the labels
inside it are set to the ignore index 255.
 
\subsubsection{Label Smoothing}
The cross-entropy target assigns $1-\epsilon$ to the ground-truth class and
$\epsilon/(C-1)$ to each of the other classes, where $C$ is the number of
classes. Pixels with the ignore index are excluded from the loss.
 
\subsubsection{NSegment}
Following~\cite{kim2025nsegment}, only the label map is deformed. Per-pixel
displacements $(\Delta x,\Delta y)\sim\mathcal{U}(-\alpha,\alpha)$ are
smoothed by a Gaussian filter with standard deviation $\sigma$, and the label
map is resampled with nearest-neighbor interpolation. For each sample,
$(\alpha,\sigma)$ is drawn uniformly from the 15 combinations in
Table~\ref{tab:supp_aug}.
 
\subsection{Shadow Augmentation Methods}
 
Table~\ref{tab:supp_shadow} lists the settings of the shadow augmentation
methods in Table~III of the main manuscript and Sec.~III. All of them modify
only the image and keep the label unchanged. Each method is applied when
$u\le p$ with $u\sim\mathcal{U}(0,1)$, and the input image is returned
unchanged when no shadow region is produced, e.g., when an image contains no
elevated-object pixel for \textbf{\textit{ShadowCast}} and SDAA. Zhong et al.\
and SynShadow resample their shadow shapes up to the number of trials in
Table~\ref{tab:supp_shadow}, so that their effective application rate remains
close to $p$.
 
\begin{table}[!htbp]
\centering
\caption{Settings of the shadow augmentation methods}
\label{tab:supp_shadow}
\footnotesize
\begin{tabular}{@{}llp{0.6\linewidth}@{}}
\toprule
Method & Implementation & Parameters \\
\midrule
\textbf{\textit{ShadowCast}} (Ours)
  & -
  & \raggedright $p=0.5$; $(d_{\min},d_{\max})=(5,30)$ pixels;
    $(\delta_{\min},\delta_{\max})=(0.5,0.8)$; $\eta=15^\circ$; $\tau=50$
    pixels; $r=8$ pixels; $\mathcal{E}=\{\text{Building},\text{Tree}\}$ for
    Vaihingen and Potsdam and $\{\text{Building},\text{Forest}\}$ for LoveDA \tabularnewline
RandomShadow~\cite{buslaev2020albumentations}
& Albumentations
& \raggedright $p=0.5$; 1 or 2 triangles with vertices sampled over the whole image;
shadow intensity $\mathcal{U}(0.5,0.8)$ \tabularnewline
PlasmaShadow~\cite{buslaev2020albumentations}
& Albumentations
& \raggedright $p=0.5$; roughness 3.0; plasma size 256;
shadow intensity $\mathcal{U}(0.5,0.8)$ \tabularnewline

SynShadow~\cite{SynShadow_inoue2020learning}
  & Ported
  & \raggedright $p=0.5$; 10\,000 rendered mattes of~\cite{SynShadow_inoue2020learning};
    $l_1\sim\mathcal{U}(0,0.25)$; $s_1\sim\mathcal{U}(0.1,0.9)$; channel
    offsets $\sim\mathcal{N}(0.05,0.025^2)$; random horizontal flip; minimum
    matte coverage 0.005; up to 5 trials \tabularnewline
SDAA~\cite{SDAA_mohajerani2021cloud}
  & Re-implemented
  & \raggedright $p=0.5$; mask $\mathbf{M}$ from $\mathcal{E}$ as in
    \textbf{\textit{ShadowCast}}; direction $\sim\mathcal{U}(0^\circ,360^\circ)$
    for each image; shift $\sim\mathcal{U}(5,30)$ pixels; gamma
    $\gamma\in\{0.800,0.825,\ldots,0.975\}$ \tabularnewline
Zhong et al.~\cite{Zhong_zhong2022shadows}
  & Ported
  & \raggedright $p=0.5$; one triangle with vertices $\sim\mathcal{U}(-0.5,1.5)$ in units
    of the image height and width; lightness scaled by
    $k\sim\mathcal{U}(0.2,0.7)$; 3-pixel edge blur; up to 10 trials \tabularnewline
SatelliteCloudGenerator~\cite{Cloud_czerkawski2023satellitecloudgenerator}
  & Ported
  & \raggedright $p=0.5$; shadow branch only; shadow level $a\sim\mathcal{U}(0.3,0.6)$;
    clear threshold 0.4; octave decay factor 1.5; constant scale \tabularnewline
\bottomrule
\multicolumn{3}{@{}l}{Ported: imported from the official code;
 \quad Re-implemented: re-implemented from the paper (no official code).}
\end{tabular}
\end{table}
 
\subsubsection{\textbf{\textit{ShadowCast}}}
Unless otherwise noted, \textbf{\textit{ShadowCast}} uses the default setting
in Table~\ref{tab:supp_shadow}. Directions are defined in image coordinates,
where $0^\circ$ points to the right ($+x$) and $90^\circ$ points downward
($+y$); the eight candidate directions in Eq.~(3) of the main manuscript are
thus spaced by $45^\circ$ clockwise from the $+x$ axis. A candidate direction
is valid when its probe region contains more than $\tau=50$ pixels, and the
darkest valid direction is always adopted without a confidence check. For each
augmented image, the jittered direction $\theta^\star$, the shift $d$, and the
attenuation factor $\delta$ are sampled once and shared by all elevated
objects, so all synthetic shadows in an image have the same direction, length,
and darkness. The confidence-gated variant evaluated in
Table~\ref{tab:supp_gate} is described in Sec.~IV-B1.
 
\subsubsection{SynShadow}
We port the affine darkening model and the shadow compositing of the official
implementation~\cite{SynShadow_inoue2020learning}. With $l_1$ and $s_1$ sampled as in
Table~\ref{tab:supp_shadow}, the slope $a=s_1/(1-l_1)$ is resampled until
$a<1$, and the channel intercepts are $l_R=l_1+\zeta_R$, $l_G=l_1$, and
$l_B=l_1-\zeta_B$ with $\zeta_R,\zeta_B\sim\mathcal{N}(0.05,0.025^2)$. For
intensities $\mathbf{x}\in[0,1]$, the shadowed image is
$m\,\mathrm{clip}(a(\mathbf{x}-\mathbf{l}),0,1)+(1-m)\,\mathbf{x}$, where
$m\in[0,1]$ is a rendered soft matte. Each matte is resized with
nearest-neighbor interpolation to cover the image, randomly cropped and
horizontally flipped, and redrawn when its mean is below 0.005.
 
\subsubsection{SDAA}
Since no official code is available, we re-implement SDAA
from~\cite{SDAA_mohajerani2021cloud}. Instead of the solar azimuth plus the
offsets $\{90^\circ,180^\circ,270^\circ\}$ of the original method, the
direction is sampled uniformly from $[0^\circ,360^\circ)$ for each image
(Sec.~\ref{sec:supp_baselines}), and the elevated-object mask $\mathbf{M}$ of
\textbf{\textit{ShadowCast}} is translated along it; only the translated pixels
outside $\mathbf{M}$ are darkened. The shift is sampled from the same range as
\textbf{\textit{ShadowCast}}, $\mathcal{U}(5,30)$ pixels, instead of the
original $\{20,40,60,80,100\}$ pixels chosen for cloud shadows in Landsat~8
imagery. Following the original
formulation on raw digital numbers, the gamma transform is applied directly to
the 8-bit intensities; for example, an intensity of 200 becomes 118 for
$\gamma=0.9$. The first step of the original method, which removes existing
shadows by using shadow annotations, is omitted, and the synthetic shadow mask
is not used as a label.
 
\subsubsection{Zhong et al.}
We port the random-shadow augmentation used for defensive training in the
official code of~\cite{Zhong_zhong2022shadows}. A triangle is drawn with vertices
sampled uniformly from $[-0.5,1.5]$ times the image height and width, the
CIELAB lightness channel (8-bit OpenCV representation) inside it is
multiplied by $k$, and a 3-pixel band along its contour is replaced by its
$3\times3$ Gaussian-blurred version. Unlike the original, which casts shadows
only inside a traffic-sign mask, the triangle can cover any part of the
image, and it is resampled when it does not overlap the image.
 
\subsubsection{SatelliteCloudGenerator}
We port the shadow branch of~\cite{Cloud_czerkawski2023satellitecloudgenerator} to
NumPy without adding clouds. A multi-octave Perlin noise with weight $s^{1.5}$
for octave scale $s$ is normalized by its maximum magnitude, compressed by
$\tanh(4\,\cdot)$, and min-max normalized to $n\in[0,1]$. The shadow pattern
$m$ is obtained by clipping $n-0.4$ to $[0,1]$ and dividing the result by its
maximum over the image, and the image is multiplied by $1-a\,m$ in all
channels. The shadows are therefore placed independently of the scene
content. With identical random numbers, the port reproduces the output of the
official PyTorch implementation.

\section{Effects across Elevated and Non-Elevated Classes of \textbf{\textit{ShadowCast}}}
\label{sec:supp_groups}

Table~\ref{tab5} reports the class-wise results grouped into elevated classes, i.e., the classes in $\mathcal{E}$ (building and tree on Vaihingen and Potsdam; building and forest on LoveDA), and non-elevated classes. \textbf{\textit{ShadowCast}} improves every class in both groups on all three datasets. The average gain is larger for the non-elevated classes on Vaihingen ($+1.11$ vs.\ $+0.68$) and LoveDA ($+1.90$ vs.\ $+0.86$), and comparable between the two groups on Potsdam ($+0.37$ vs.\ $+0.38$).

\begin{table}[ht!]
\centering
\caption{Per-group/Per-class mIoU of \textit{\textbf{ShadowCast}} (DeepLabV3+).}  \vspace{-0.2cm}
\label{tab5}
\resizebox{0.8\columnwidth}{!}{
\setlength{\tabcolsep}{1.5pt}
\renewcommand{\arraystretch}{1.05}
\begin{tabular}{l cc cc @{\hspace{2em}} l cc}
\toprule
\multirow{2}{*}{Class} & \multicolumn{2}{c}{Vaihingen} & \multicolumn{2}{c}{Potsdam} &
\multirow{2}{*}{Class} & \multicolumn{2}{c}{LoveDA} \\
\cmidrule(lr){2-3} \cmidrule(lr){4-5} \cmidrule(lr){7-8}
 & Baseline & \cellcolor[HTML]{FFCE93}\textit{\textbf{ShadowCast}}
 & Baseline & \cellcolor[HTML]{FFCE93}\textit{\textbf{ShadowCast}}
 & & Baseline & \cellcolor[HTML]{FFCE93}\textit{\textbf{ShadowCast}} \\
\midrule
\multicolumn{5}{l}{\emph{\textbf{Non-elevated classes}}} & \multicolumn{3}{l}{ } \\
Impervious surface & 83.08$\pm$0.17 & \cellcolor[HTML]{FFCE93}\textbf{83.93$\pm$0.06 (+0.85)}
                   & 84.49$\pm$0.25 & \cellcolor[HTML]{FFCE93}\textbf{84.87$\pm$0.14 (+0.38)}
 & Background   & 49.30$\pm$0.73 & \cellcolor[HTML]{FFCE93}\textbf{50.19$\pm$0.12 (+0.89)} \\
Low vegetation     & 68.44$\pm$0.55 & \cellcolor[HTML]{FFCE93}\textbf{69.36$\pm$0.60 (+0.92)}
                   & 73.50$\pm$0.25 & \cellcolor[HTML]{FFCE93}\textbf{73.82$\pm$0.05 (+0.32)}
 & Road         & 48.40$\pm$0.96 & \cellcolor[HTML]{FFCE93}\textbf{49.24$\pm$0.71 (+0.84)} \\
Car                & 69.48$\pm$0.61 & \cellcolor[HTML]{FFCE93}\textbf{71.05$\pm$0.19 (+1.57)}
                   & 89.94$\pm$0.24 & \cellcolor[HTML]{FFCE93}\textbf{90.35$\pm$0.26 (+0.41)}
 & Water        & 47.89$\pm$1.71 & \cellcolor[HTML]{FFCE93}\textbf{54.39$\pm$0.97 (+6.50)} \\
 & & \cellcolor[HTML]{FFCE93} & & \cellcolor[HTML]{FFCE93}
 & Barren       & 19.48$\pm$0.96 & \cellcolor[HTML]{FFCE93}\textbf{20.38$\pm$1.45 (+0.90)} \\
 & & \cellcolor[HTML]{FFCE93} & & \cellcolor[HTML]{FFCE93}
 & Agricultural & 47.49$\pm$1.33 & \cellcolor[HTML]{FFCE93}\textbf{47.86$\pm$0.57 (+0.37)} \\
\multicolumn{2}{>{\cellcolor[HTML]{FFCE93}}l}{\textit{Mean mIoU improvement}}
 & \cellcolor[HTML]{FFCE93}{\textbf{+1.11}}
 & \cellcolor[HTML]{FFCE93}
 & \cellcolor[HTML]{FFCE93}{\textbf{+0.37}}
 & \cellcolor[HTML]{FFCE93}
 & \cellcolor[HTML]{FFCE93}
 & \cellcolor[HTML]{FFCE93}{\textbf{+1.90}} \\ \midrule\midrule
\addlinespace
\multicolumn{5}{l}{\emph{\textbf{Elevated classes}}} & \multicolumn{3}{l}{} \\
Building           & 88.62$\pm$0.28 & \cellcolor[HTML]{FFCE93}\textbf{89.48$\pm$0.17 (+0.86)}
                   & 91.37$\pm$0.20 & \cellcolor[HTML]{FFCE93}\textbf{91.83$\pm$0.30 (+0.46)}
 & Building     & 51.81$\pm$0.74 & \cellcolor[HTML]{FFCE93}\textbf{52.72$\pm$0.80 (+0.91)} \\
Tree               & 78.01$\pm$0.50 & \cellcolor[HTML]{FFCE93}\textbf{78.50$\pm$0.42 (+0.49)}
                   & 75.42$\pm$0.17 & \cellcolor[HTML]{FFCE93}\textbf{75.72$\pm$0.09 (+0.30)}
 & Forest       & 38.63$\pm$1.05 & \cellcolor[HTML]{FFCE93}\textbf{39.43$\pm$0.20 (+0.80)} \\
\multicolumn{2}{>{\cellcolor[HTML]{FFCE93}}l}{\textit{Mean mIoU improvement}}
 & \cellcolor[HTML]{FFCE93}{\textbf{+0.68}}
 & \cellcolor[HTML]{FFCE93}
 & \cellcolor[HTML]{FFCE93}{\textbf{+0.38}}
 & \cellcolor[HTML]{FFCE93}
 & \cellcolor[HTML]{FFCE93}
 & \cellcolor[HTML]{FFCE93}{\textbf{+0.86}} \\
\bottomrule
\end{tabular}
}
\end{table}

\subsection{Non-Elevated Classes}

In the default augmentation, the projected mask excludes all pixels in
$\mathcal{E}$. 
    Synthetic shadows therefore usually change the appearance of non-elevated classes while preserving their labels. 
These classes are directly exposed to variants of brightness during training, providing a straightforward explanation for their improvement.
    This observation is also consistent with the gains on regions covered by real cast shadows reported in the shadow-region evaluation.

\subsection{Elevated Classes}

Elevated classes also improve, although the default augmentation never
directly darkens their pixels in most cases. Two possible explanations are consistent with this observation.
    First, learning to recognize non-elevated classes under brightness changes may encourage features that depend less on absolute brightness. Such features could also help recognize buildings and trees covered by real shadows. 
Examples include the shaded tree in Fig.~\ref{fig:supp_fail}(b) and the buildings shaded by neighboring high-rise blocks in Fig.~\ref{fig:supp_fail}(c).
    Second, synthetic shadows often alter the appearance of regions near elevated-object boundaries.
Better discrimination of the surrounding classes could consequently benefit the delineation of the adjacent elevated classes.


\section{Design Choices and Additional Ablation Studies of \textbf{\textit{ShadowCast}}}
\label{supp:sec_shadow_design}

\textbf{\textit{ShadowCast}} is designed for stochastic augmentation
during segmentation training. It derives potential casters from
existing labels, estimates an image-specific direction from local
brightness statistics, and samples the displacement, angular
perturbation, and attenuation. This permits on-the-fly augmentation
without shadow annotations, additional learned models, acquisition
metadata, or pre-generated augmented images. The formulation uses rigid
mask translation and uniform multiplicative attenuation, and does not
explicitly model penumbra, complex light transport, or
material-dependent reflectance.

We examine two simple modifications motivated by these limitations:
smoothing the synthetic shadow boundary and rejecting direction
estimates that fail brightness-based reliability checks. Both
modifications yield lower mean mIoU than the default configuration in
the tested settings. We report these outcomes to distinguish the
motivation for a modification from its observed effect on segmentation.

\subsection{Hard versus Soft Shadow Boundaries}
\label{sec:supp_blur}

The default formulation applies the same multiplicative attenuation to
every pixel in the binary shadow region, producing a hard boundary
(Eq.~(9) of the main manuscript). To test a simple soft-boundary
alternative, we smooth the binary mask with a Gaussian kernel and use
the result as an alpha map for blending the original and attenuated
images. For blur width $b$, the kernel size is
$(2b+1)\times(2b+1)$. We compare the default hard mask ($b=0$) with
$b=1$, keeping the direction-estimation procedure, displacement and
attenuation sampling, and training schedule unchanged.

\begin{table}[h]
\centering
\caption{mIoU (\%) of DeepLabV3+ with Hard ($b{=}0$) and Soft ($b{=}1$) 
Synthetic Shadows in \textbf{\textit{ShadowCast}}.}
\label{tab:supp_blur}
\resizebox{0.6\columnwidth}{!}{
\setlength{\tabcolsep}{1.5pt}
\renewcommand{\arraystretch}{1.05}
\begin{tabular}{lccc}
\hline
 & Vaihingen & Potsdam & LoveDA \\
\hline
Baseline                      & 77.53$\pm$0.30 & 82.95$\pm$0.10 & 43.29$\pm$0.77 \\
\cellcolor[HTML]{FFCE93}\textbf{\textit{ShadowCast}} (Default: Hard-shadow) ($b{=}0$) & \cellcolor[HTML]{FFCE93}\textbf{78.46$\pm$0.19} & \cellcolor[HTML]{FFCE93}\textbf{83.32$\pm$0.09} & \cellcolor[HTML]{FFCE93}\textbf{44.89$\pm$0.09} \\
\cellcolor[HTML]{FFCE93}\textbf{\textit{ShadowCast}} (Soft-shadow) ($b{=}1$)         & \cellcolor[HTML]{FFCE93}77.79$\pm$0.12 & \cellcolor[HTML]{FFCE93}82.98$\pm$0.13 & \cellcolor[HTML]{FFCE93}43.36$\pm$0.24 \\
\hline
\end{tabular}
}
\end{table}

\begin{figure}[!htb]
\centering
\includegraphics[width=\linewidth]{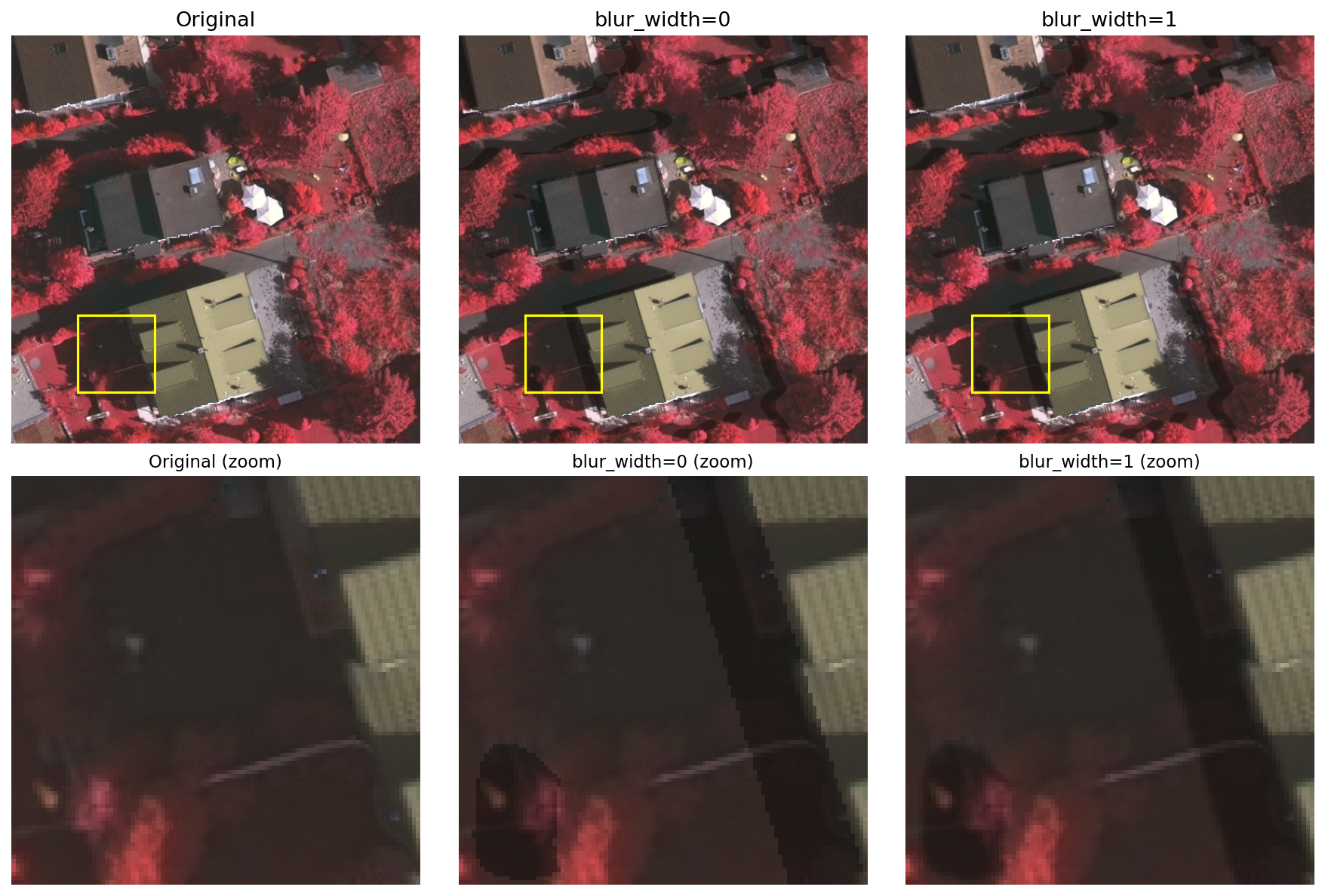}
\caption{Hard ($b=0$) and soft ($b=1$) synthetic shadow boundaries on
a Vaihingen patch. The bottom row enlarges the region marked by the
\colorbox{black}{\textcolor{yellow}{\textbf{yellow}}} box. }
\label{fig:supp_blur}
\end{figure}

As shown in Table~\ref{tab:supp_blur}, the soft variant produces mean
scores only slightly above the baseline. Its gains are $+0.26$,
$+0.03$, and $+0.07$ on Vaihingen, Potsdam, and LoveDA, compared with
$+0.93$, $+0.37$, and $+1.60$ for the default hard mask.
Thus, this simple smoothing modification does not preserve most of the
segmentation benefit of the default formulation.

Fig.~\ref{fig:supp_blur} illustrates how smoothing redistributes
attenuation around the boundary: the transition becomes gradual, and
pixels just inside the original boundary are less strongly darkened.
The comparison therefore changes both boundary sharpness and the
spatial distribution of attenuation. It does not isolate the effect of
a soft boundary independently of the resulting appearance change.
We retain hard masks based on the reported segmentation results, but
these results do not establish that soft shadows are generally less
useful. A formulation that improves boundary realism while preserving
the beneficial appearance variation requires further investigation.

\subsection{Confidence-Gated Direction Estimation}
\label{sec:supp_gate}

The default estimator always selects the darkest valid direction, even
when the directional contrast is weak or the dark region has a
different origin. To test whether rejecting ambiguous estimates is
beneficial, we add a gap check and an opposite-side brightness check in \textbf{\textit{ShadowCast}}.

Let $\mu_\theta=\mathrm{mean}(\mathbf{I}'[\mathbf{R}_\theta])$ for
$\theta\in\Theta'$, and let
\[
\tilde{\theta}=\arg\min_{\theta\in\Theta'}\mu_\theta,
\qquad
\mu_{\mathrm{ref}}=
\frac{1}{|\Theta'|}\sum_{\theta\in\Theta'}\mu_\theta.
\]
For the gap check, we exclude $\tilde{\theta}$ and its two adjacent
directions, with angles interpreted modulo $2\pi$:
\[
\Theta_{\mathrm{sep}}=
\Theta'\setminus
\{\tilde{\theta}-\pi/4,\tilde{\theta},\tilde{\theta}+\pi/4\},
\qquad
\mu_{(2)}=\min_{\theta\in\Theta_{\mathrm{sep}}}\mu_\theta.
\]
The two scores are
\[
\mathrm{gap}=
\frac{\mu_{(2)}-\mu_{\tilde{\theta}}}{\mu_{\mathrm{ref}}},
\qquad
\mathrm{sym}=
\frac{\mu_{\tilde{\theta}+\pi}-\mu_{\tilde{\theta}}}
{\mu_{\mathrm{ref}}}.
\]
We require
$\mathrm{gap}\ge\tau_{\mathrm{gap}}$ and
$\mathrm{sym}\ge\tau_{\mathrm{sym}}$.
Adjacent directions are excluded from the gap check because a real
shadow can make several nearby directions dark. 
    The second check requires the opposite side to be brighter, providing an additional heuristic cue for a directional shadow. 
Normalization by $\mu_{\mathrm{ref}}$ reduces dependence on overall brightness.

If either check fails or cannot be evaluated, the image is returned
unchanged. In particular, the checks fail when
$\Theta_{\mathrm{sep}}=\emptyset$ or
$\tilde{\theta}+\pi\notin\Theta'$; the original skip rule also applies
when $\Theta'=\emptyset$.
We set $\tau_{\mathrm{gap}}=0.03$ and
$\tau_{\mathrm{sym}}=0.10$, and retain all other default parameters in
Table~\ref{tab:supp_shadow}. The angular jitter in Eq.~(7) of the main
manuscript is applied only after both checks pass.

\begin{table}[!htbp]
\centering
\caption{Effect of the confidence-gated direction estimation
($\tau_{\mathrm{gap}}=0.03$, $\tau_{\mathrm{sym}}=0.10$) on DeepLabV3+ for the
Vaihingen, Potsdam, and LoveDA benchmarks (mIoU, \%)}
\label{tab:supp_gate}
\resizebox{0.7\columnwidth}{!}{
\setlength{\tabcolsep}{1.5pt}
\renewcommand{\arraystretch}{1.05}
\begin{tabular}{@{}lccc@{}}
\toprule
Method & Vaihingen & Potsdam & LoveDA \\
\midrule
Baseline
  & 77.53$\pm$0.30 & 82.95$\pm$0.10 & 43.29$\pm$0.77 \\
\cellcolor[HTML]{FFCE93}\textbf{\textit{ShadowCast}} (Default)
  & \cellcolor[HTML]{FFCE93}\textbf{78.46$\pm$0.19 ($+$0.93)} & \cellcolor[HTML]{FFCE93}\textbf{83.32$\pm$0.09 ($+$0.37)}
  & \cellcolor[HTML]{FFCE93}\textbf{44.89$\pm$0.09 ($+$1.60)} \\
\cellcolor[HTML]{FFCE93}\textbf{\textit{ShadowCast}} (+ Confidence checks)
  & \cellcolor[HTML]{FFCE93}78.18$\pm$0.57 ($+$0.65) & \cellcolor[HTML]{FFCE93}83.22$\pm$0.04 ($+$0.27)
  & \cellcolor[HTML]{FFCE93}43.90$\pm$0.30 ($+$0.61) \\
\bottomrule
\end{tabular}
}
\end{table}

Table~\ref{tab:supp_gate} shows that the gated variant still improves
over the baseline on all three datasets, but falls below the default
by $0.28$, $0.10$, and $0.99$ mIoU points on Vaihingen, Potsdam, and
LoveDA, respectively. These checks therefore do not improve mean
segmentation performance at the tested thresholds.
    Rejecting images reduces the frequency of shadow augmentation, which may contribute to the lower scores. 
We therefore do not use the confidence-gate check in our method for remaining experiments. 
    Further work should evaluate direction reliability separately from augmentation frequency to determine whether a more selective strategy can improve sample quality without losing the segmentation benefit.



\section{Failure Cases and Limitations}
\label{sec:supp_failure}

\begin{figure}[!htb]
\centering
\setlength{\tabcolsep}{3pt}
\begin{tabular}{ccc}
\includegraphics[width=0.30\linewidth]{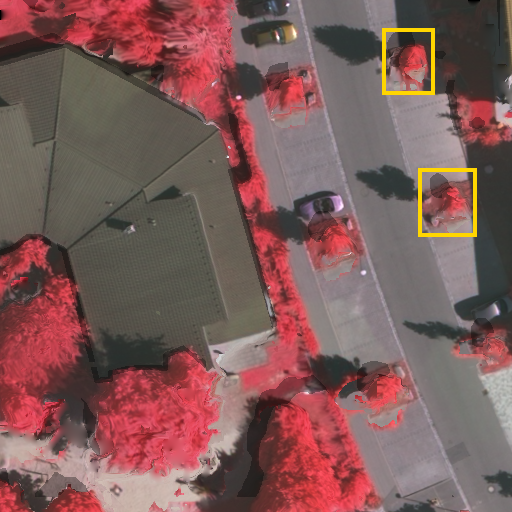} &
\includegraphics[width=0.30\linewidth]{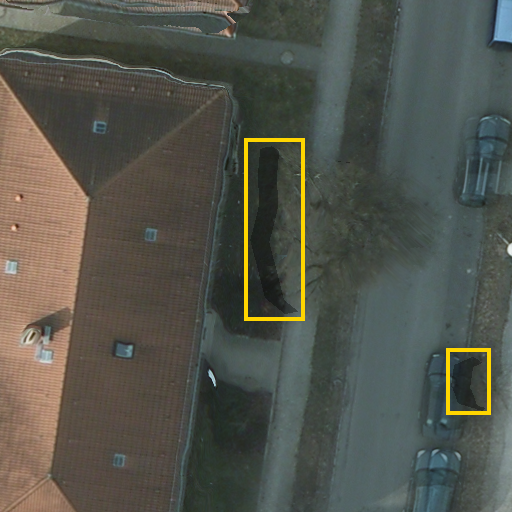} &
\includegraphics[width=0.30\linewidth]{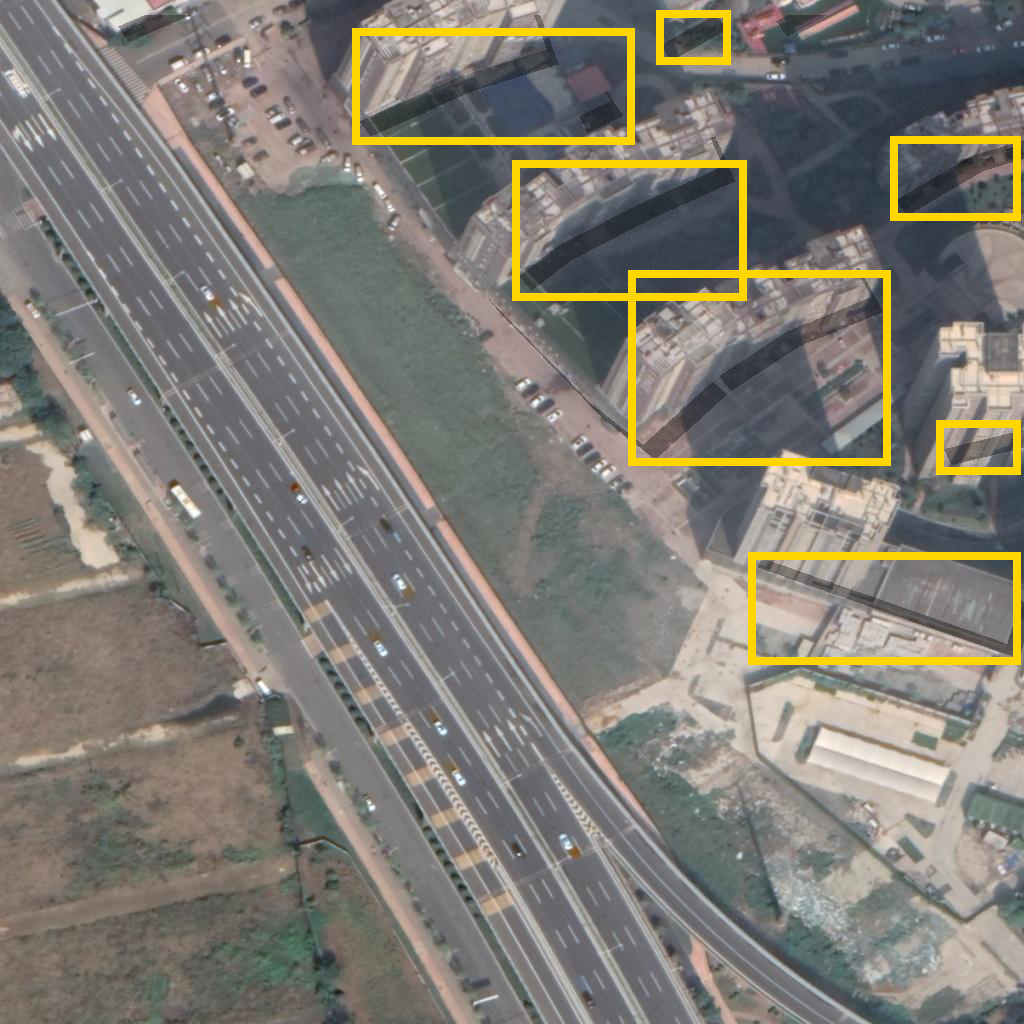} \\
{\footnotesize (a) Vaihingen} &
{\footnotesize (b) Potsdam} &
{\footnotesize (c) LoveDA}
\end{tabular}
\caption{Representative failure cases of
\textbf{\textit{ShadowCast}} on training images from
(a) Vaihingen, (b) Potsdam, and (c) LoveDA; see \colorbox{black}{\textcolor[RGB]{255,214,0}{\textbf{Gold}}} bounding boxes. The examples show incorrect shadow length and shape, an unclear direction in an already shaded area, and misleading shadows from neighboring objects,
respectively.}
\label{fig:supp_fail}
\end{figure}

The experiments in Sections~\ref{sec:supp_blur} and \ref{sec:supp_gate} show that simply smoothing shadow boundaries or rejecting estimates with weak brightness cues does not outperform the default configuration in the tested settings.
We now examine the underlying limitations in more detail.
\textbf{\textit{ShadowCast}} assumes that the brightness around labeled elevated objects indicates the dominant shadow direction, and that shifting the object masks approximates their shadows on locally flat ground.
Fig.~\ref{fig:supp_fail} shows three typical failures: incorrect shadow length and shape, an unclear direction in already shaded areas, and misleading shadows from neighboring objects. Table~\ref{tab:supp_fail_taxonomy} summarizes their causes.
Possible extensions are discussed alongside the corresponding failure.
    Apart from the variants evaluated in Sec.~\ref{supp:sec_shadow_design}, these potential extensions have not been evaluated and are left for future work.

\begin{table}[!htb]
\caption{Categorization of the Failure Cases in Fig.~\ref{fig:supp_fail}}
\label{tab:supp_fail_taxonomy}
\centering
\footnotesize
\renewcommand{\arraystretch}{1.15}
\begin{tabular}{l l p{0.27\linewidth} p{0.33\linewidth}}
\hline
Limitation & Example & Observed failure & Underlying cause \\
\hline
Shadow length and shape & Fig.~\ref{fig:supp_fail}(a) & Shadows too short; shape does not follow the object & Same shift range for all object heights; single shift of the top-view mask \\
Unclear direction & Fig.~\ref{fig:supp_fail}(b) & Arbitrary direction in an already shaded area & No brightness difference between sides \\
Neighboring shadows & Fig.~\ref{fig:supp_fail}(c) & Wrong side appears darkest & No link between shadows and casters; averaging over all objects \\
\hline
\end{tabular}
\end{table}

\subsection{Observed Failure Cases}

\subsubsection{Incorrect Shadow Length and Shape}

In Fig.~\ref{fig:supp_fail}(a), the estimated direction is correct: the synthetic shadows point toward the upper left, as the real shadows do.
The problem is their length and shape. 
    The synthetic shadows remain close to the trees and fail to reproduce the longer, displaced shadows observed on the road.
        This happens because \textbf{\textit{ShadowCast}} moves every object mask by a randomly chosen distance that does not depend on the height of the object. 
    In reality, taller objects cast longer shadows, so a single distance range cannot fit both low and tall objects. 
In addition, the mask is moved only once, so the synthetic shadow is simply a shifted copy of the object as seen from above. 
        It cannot reproduce shapes that depend on the 3D structure of the object, such as the shadow of a tree that appears separated from the tree because its leaves are held high above the ground by the trunk.
    A possible remedy is to incorporate height information, for example from a digital surface model, to better account for object-dependent shadow lengths and shapes.
        However, using height maps together with the images and labels, along with the heavier shadow computation they involve, would require a larger cost.

\subsubsection{Unclear Direction in Already Shaded Areas}

In Fig.~\ref{fig:supp_fail}(b), the tree stands in an area that is already largely in shadow. \textbf{\textit{ShadowCast}} estimates the direction by finding the side of the objects that is darkest on average. 
    Here, however, all sides are similarly dark, so the brightness no longer indicates a clear direction.
\textbf{\textit{ShadowCast}} must nevertheless choose one, and small variations in local texture effectively decide the result.
    A possible remedy is to improve the direction estimate by identifying and down-weighting ambiguous regions affected by neighboring shadows, rather than relying on uniformly averaged brightness measurements.

\subsubsection{Misleading Shadows from Neighboring Objects}

Fig.~\ref{fig:supp_fail}(c) shows high-rise apartment buildings arranged in closely spaced rows.
        \textbf{\textit{ShadowCast}} estimates the shadow direction from the brightness around each building, but cannot distinguish shadows cast by the target building from those cast by neighboring buildings. 
In such dense layouts, a neighboring building's shadow may cause the wrong side of a building to appear darkest. 
    Because similar building arrangements recur across the image, these biased measurements can reinforce one another when averaged to estimate a single global direction. 
In addition, the visible side walls of the buildings lie adjacent to the roof masks, further biasing the brightness measurements.
    A possible remedy is to identify and down-weight unreliable object-level measurements when estimating the global shadow direction, rather than treating all objects equally.

\subsection{Terrain with Large Height Variation}

The benchmarks used in this work mainly contain shadows of buildings and
trees on relatively flat ground, so they do not show whether
\textbf{\textit{ShadowCast}} works in hilly or mountainous areas.
\textbf{\textit{ShadowCast}} assumes that the ground around each object is
flat and that only the labeled elevated objects cast shadows. In terrain
with large height differences, these assumptions break down in several ways:
\begin{itemize}
\item \textit{Shadows change on slopes.}
On sloped ground, the length and shape of a shadow depend on the slope.
Shadows become longer on slopes facing away from the sun, shorter on slopes
facing toward it, and bend where the slope changes. Shifting a mask by a
fixed distance cannot reproduce these changes.
\item \textit{Shadows cast by the terrain itself.}
Hills and steep slopes can cast their own shadows~\cite{wang2023shadow}, and slopes facing away
from the sun can be dark even without any object nearby. These shadows are
not linked to any labeled object, so the augmentation does not
model them.
\item \textit{Darkness caused by slope, not by shadow.}
The brightness of the ground also depends on which way it faces. A dark
probe region may therefore reflect the slope rather than a cast shadow,
which can mislead the direction estimate in the same way as in
Fig.~\ref{fig:supp_fail}(b).
\item \textit{Tall objects leaning in the image.}
In images that are not corrected for relief displacement, tall objects appear to
lean away from their actual ground position, as with the visible building
walls in Fig.~\ref{fig:supp_fail}(c). The object masks then no longer match
where the shadows start on the ground.
\end{itemize}

If a height map of the scene and the sun position are available, shadows
can be computed physically for the actual terrain, as done for the shadow
masks in S-EO~\cite{masquil2025s}. However, this requires inputs that
\textbf{\textit{ShadowCast}} does not use and a different way of placing
shadows. 
    Since \textbf{\textit{ShadowCast}} has not been evaluated on terrain with substantial elevation variations, its effectiveness under such conditions remains uncertain.

Nevertheless, the gains in Table~I of the main manuscript are obtained even though the synthetic shadows ignore object height, as shown in Fig.~\ref{fig:supp_fail}(a). 
    These results suggest that precise shadow geometry is not always necessary to improve segmentation performance on the evaluated datasets. 
However, whether this benefit extends to complex terrain remains to be investigated.

\end{document}